\documentclass[twoside,11pt]{article}

\usepackage[accepted]{melba}

\usepackage{mwe} 

\usepackage{amsfonts}

\usepackage{amssymb}
\usepackage{latexsym}
\usepackage{url}
\usepackage{xcolor}

\usepackage{amsmath}
\usepackage{xspace}
\usepackage{hyperref}
\usepackage{subfig}
\usepackage{amssymb}
\usepackage{nicefrac}
\usepackage{booktabs}
\usepackage{microtype}
\usepackage[inline]{enumitem}
\usepackage{multirow}
\usepackage{bm}
\usepackage{soul,color}
\usepackage{placeins}
\usepackage[utf8]{inputenc}
\usepackage[T1]{fontenc}
\usepackage{mathtools}
\usepackage{comment}
\usepackage[english]{babel}
\usepackage{blindtext}
\usepackage{breakcites}
\usepackage{pbox}
\usepackage{makecell}
\usepackage{diagbox}
\usepackage{tablefootnote}
\usepackage{colortbl}
\usepackage{pifont}

\DeclareMathAlphabet{\mathalphm}{OMS}{cmsy}{m}{n}
\DeclareMathAlphabet{\mathalphb}{OMS}{cmsy}{b}{n}

\newcommand{\R}{\mathbb{R} \xspace}

\newcommand{\Y}{\mathalphm{Y} \xspace}

\DeclareMathOperator*{\Prob}{Pr}

\DeclareMathOperator{\softmax}{softmax}

\DeclareMathOperator{\relu}{ReLU}

\makeatletter
\DeclareRobustCommand\onedot{\futurelet\@let@token\@onedot}
\def\@onedot{\ifx\@let@token.\else.\null\fi\xspace}

\def\eg{e.g\onedot} 

\def\ie{i.e\onedot}

\makeatother

\DeclareMathOperator*{\argmax}{argmax}

\definecolor{firebrick}{rgb}{.698,.133,.133}
\definecolor{lightg}{gray}{.9}

\newcommand{\midsepremove}{\aboverulesep = 0mm \belowrulesep = 0mm}
\newcommand{\midsepdefault}{\aboverulesep = 0.605mm \belowrulesep = 0.984mm}
\newcolumntype{g}{>{\columncolor{lightg}}c}
\newcommand\pxap{\texttt{PxAP}\xspace}
\newcommand\pxprec{\texttt{PxPrec}\xspace}
\newcommand\pxrec{\texttt{PxRec}\xspace}

\newcommand\tp{\texttt{TP}\xspace}
\newcommand\tn{\texttt{TN}\xspace}
\newcommand\fp{\texttt{FP}\xspace}
\newcommand\fn{\texttt{FN}\xspace}

\newcommand\pxtp{\texttt{TP*}\xspace}
\newcommand\pxtn{\texttt{TN*}\xspace}
\newcommand\pxfp{\texttt{FP*}\xspace}
\newcommand\pxfn{\texttt{FN*}\xspace}

\newcommand\cmtx{\texttt{Confusion Matrix}\xspace}
\newcommand\cl{\texttt{CL}\xspace}

\newcommand\glas{\texttt{GlaS}\xspace}
\newcommand\camsixteen{\texttt{CAMELYON16}\xspace}

\newcommand\beloc{\texttt{B-LOC}\xspace}
\newcommand\becl{\texttt{B-CL}\xspace}

\definecolor{darkergreen}{RGB}{21, 152, 56}
\definecolor{red2}{RGB}{252, 54, 65}
\definecolor{Gray}{gray}{0.85}

\newcommand\red[1]{\textcolor{red2}{#1}}
\newcommand\green[1]{\textcolor{darkergreen}{#1}}

\melbaid{2023:004}  
\melbaauthors{Rony et al.}
\volume{2}
\firstpageno{96}  
\year{2023}  
\datesubmitted{07/2022}  
\datepublished{03/2023}  

\ShortHeadings{Deep Weakly-Supervised Learning for Histology Images: A Survey}{Rony et al.}

\title{Deep Weakly-Supervised Learning Methods for Classification and Localization in Histology Images: A Survey}

\author{\name Jérôme Rony \email jerome.rony.1@etsmtl.net \\  
	\addr LIVIA, Dept. of Systems Engineering, École de technologie supérieure, Montreal, Canada
	\AND
	\name Soufiane Belharbi \email soufiane.belharbi.1@ens.etsmtl.ca \\
	\addr LIVIA, Dept. of Systems Engineering, École de technologie supérieure, Montreal, Canada
	\AND
	\name Jose Dolz \email jose.dolz@etsmtl.ca \\
	\addr LIVIA, Dept. of Software and IT Engineering, École de technologie supérieure, Montreal, Canada
	\AND
	\name Ismail Ben Ayed \email ismail.benayed@etsmtl.ca \\
	\addr LIVIA, Dept. of Systems Engineering, École de technologie supérieure, Montreal, Canada
	\AND
	\name Luke McCaffrey \email luke.mccaffrey@mcgill.ca \\
	\addr Goodman Cancer Research Centre, Dept. of Oncology, McGill University, Montreal, Canada
	\AND
	\name Eric Granger \email eric.granger@etsmtl.ca \\
	\addr LIVIA, Dept. of Systems Engineering, École de technologie supérieure, Montreal, Canada
}

\begin{document}

\maketitle

\begin{abstract}
Using state-of-the-art deep learning (DL) models to diagnose cancer from histology data presents several challenges related to the nature and availability of labeled histology images, including image size, stain variations, and label ambiguity. In addition, cancer grading and the localization of regions of interest (ROIs) in such images normally rely on both image- and pixel-level labels, with the latter requiring a costly annotation process. Deep weakly-supervised object localization (WSOL) methods provide different strategies for low-cost training of DL models. Given only image-class annotations, these methods can be trained to simultaneously classify an image, and yield class activation maps (CAMs) for ROI localization. 
This paper provides a review of deep WSOL methods to identify and locate diseases in histology images, without the need for pixel-level annotations. We propose a taxonomy in which these methods are divided into bottom-up and top-down methods according to the information flow in models. Although the latter have seen only limited progress, recent bottom-up methods are currently  driving a lot of progress with the use of deep WSOL methods. Early works focused on designing different spatial pooling functions. However, those methods quickly peaked in term of localization accuracy and revealed a major limitation, namely, -- the under-activation of CAMs, which leads to high false negative localization. Subsequent works aimed to alleviate this shortcoming and recover the complete object from the background, using different techniques such as perturbation, self-attention, shallow features, pseudo-annotation, and task decoupling. 
In the present paper, representative deep WSOL methods from our taxonomy are also evaluated and compared in terms of classification and localization accuracy using two challenging public histology datasets -- one for colon cancer (\glas), and a second, for breast cancer (\camsixteen).  Overall, the results indicate poor localization performance, particularly for generic methods that were initially designed to process natural images. Methods designed to address the challenges posed by histology data often use priors such as ROI size, or additional pixel-wise supervision estimated from a pre-trained classifier, allowing them to achieve better results. However, all the methods suffer from high false positive/negative localization. Classification performance is mainly affected by the model selection process, which uses either the classification or the localization metric. Finally, four key challenges are identified in the application of deep WSOL methods in histology, namely, -- under-/over-activation of CAMs, sensitivity to thresholding, and model selection -- and research avenues are provided to mitigate them.  Our code is publicly available at \url{https://github.com/jeromerony/survey\_wsl\_histology}.
\end{abstract}


\section{Introduction}
\label{sec:introduction} 

The advent of Whole Slide Imaging (WSI) scanners opened new possibilities in pathology image analysis~\citep{he2012histology,madabhushi2009digital}. Histology slides provide more comprehensive views of diseases and of their effect on tissue~\citep{hipp2011pathology}, since their preparation preserves the underlying tissue structure~\citep{he2012histology}. For instance, some disease characteristics (\eg, lymphatic infiltration of cancer) may be predicted using only histology images~\citep{gurcan2009histopathological}. Histology images analysis remains the gold standard in diagnosing several diseases, including most types of cancer~\citep{gurcan2009histopathological,he2012histology,Veta2014}. Breast cancer, which is the most prevalent cancer in women worldwide, relies on medical imaging systems as a primary diagnostic tool for its early detection~\citep{Daisuke201834,Veta2014,xie2019deep}. 

Cancer is mainly diagnosed by pathologists who analyze WSIs to identify and assess epithelial cells organized into ducts, lobules, or malignant clusters, and embedded within a heterogeneous stroma. Manual analysis of histology tissues depends heavily on the expertise and experience of histopathologists. Such manual interpretation is time-consuming and difficult to grade in a reproducible manner. Analyzing WSIs from digitized histology slides enables facilitated, and potentially automated, Computer-Aided Diagnosis in pathology, where the main goal is to confirm the presence or absence of disease and to grade or measure disease progression. 

Given the large number of digitized exams in use, automated systems have become a part of the clinical routines for breast cancer detection~\citep{tang2009computer}. Automated analysis of the spatial structures in histology images can be traced back to early works~\citep{bartels1992bayesian,hamilton1994expert,weind1998invasive}. Various image processing and machine learning (ML) techniques have been investigated in a bid to identify discriminative structures and classify histology images~\citep{he2012histology}; these include thresholding~\citep{gurcan2006image, petushi2006large}, active contours~\citep{Bamford2001}, Bayesian classifiers~\citep{Naik2007}, graphs used to model spatial structures~\citep{bilgin2007cell, tabesh2007multifeature}, and ensemble methods based on Support Vector Machines and Adaboost~\citep{doyle2006detecting, qureshi2008adaptive}. An overview of these techniques and their applications is provided in~\citep{gurcan2009histopathological,he2012histology, Veta2014}. Recently, deep learning (DL) models have attracted a lot of attention in histology image analysis~\citep{belharbi2020DeepAlJoinClSegWeakAnn,negevsbelharbi2022,belharbi2019weakly,belharbi2022minmaxuncer,courtiol2018classification,Dimitriou2019,iizuka2020deep,janowczyk2016deep,li2018cancer,srinidhi2019deep}. In the present paper, we continue in the same vein and focus on the application of DL models in histology image analysis.

DL models~\citep{Goodfellow-et-al-2016}, and convolutional neural networks (CNNs) in particular, provide state-of-the-art performance in many visual recognition applications such as image classification~\citep{KrizhevskyNIPS2012}, object detection~\citep{RedmonDGF16CVPR}, and segmentation~\citep{dolz20183d}. These supervised learning architectures are trained end-to-end with large amounts of annotated data. More recently, the potential of DL models has begun to be explored in assisted pathology diagnosis~\citep{Daisuke201834,janowczyk2016deep,li2018cancer}. Given the growing availability of histology slides, DL models have not only been proposed for disease prediction~\citep{hou2016patch,li2018cancer,sheikhzadeh2016automatic,Spanhol2016,xu2016deep}, but also for related tasks such as the detection and segmentation of tumor regions within WSI~\citep{kieffer2017convolutional,mungle2017mrf}, scoring of immunostaining~\citep{sheikhzadeh2016automatic,wang2015exploring}, cancer staging~\citep{shah2017deep,Spanhol2016}, mitosis detection \citep{chen2016dcan,cirecsan2013mitosis,roux2013mitosis}, gland segmentation~\citep{caie2014quantification,gertych2015machine,sirinukunwattana2017gland}, and detection and quantification of vascular invasion~\citep{caicedo2011content}.

\begin{figure}[!b]
    \centering
    \subfloat[]{
    \includegraphics[width=0.4\linewidth]{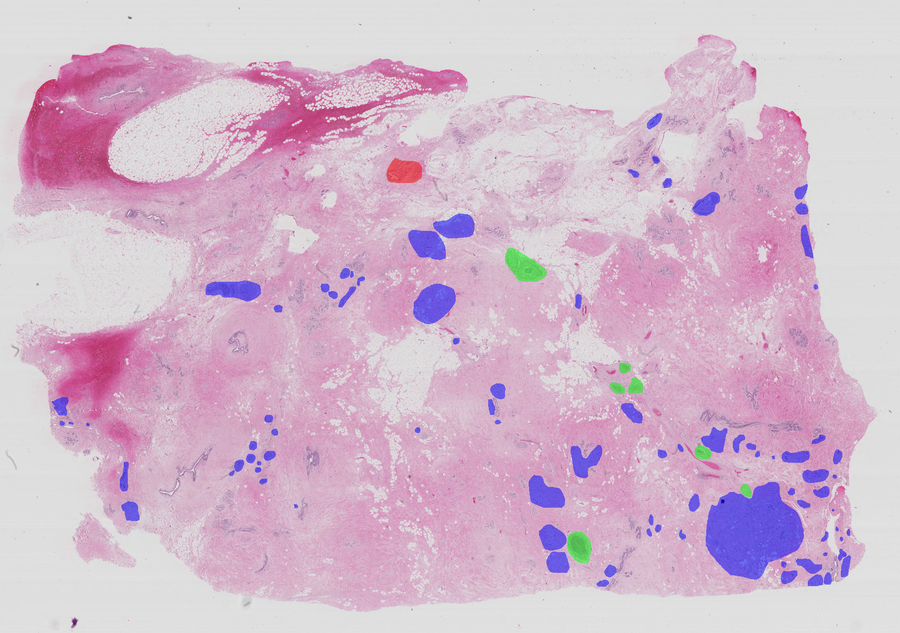}}
    \subfloat[]{
    \includegraphics[width=0.4\linewidth]{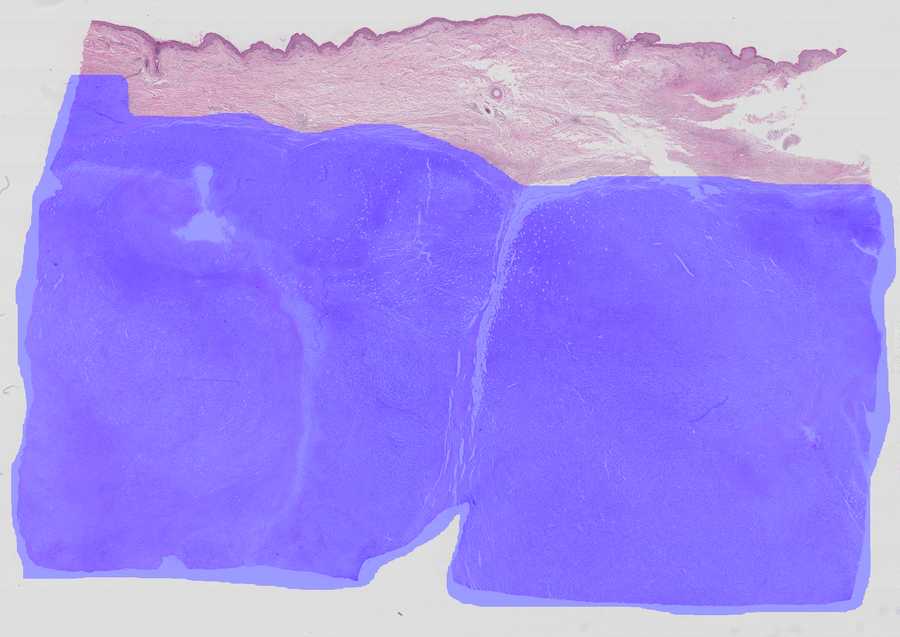}}
    \caption{Segmentation of two WSIs from the ICIAR 2018 BACH Challenge. Colors represent different types of cancerous regions: red for \texttt{Benign}, green for \texttt{In Situ Carcinoma} and blue for \texttt{Invasive Carcinoma}. These examples highlight the diversity in size and regions~\citep{aresta2018bach}.}
    \label{fig:wsisegmentationannot}
\end{figure}

\begin{figure}[!t]
  \centering
  \includegraphics[width=0.8\linewidth]{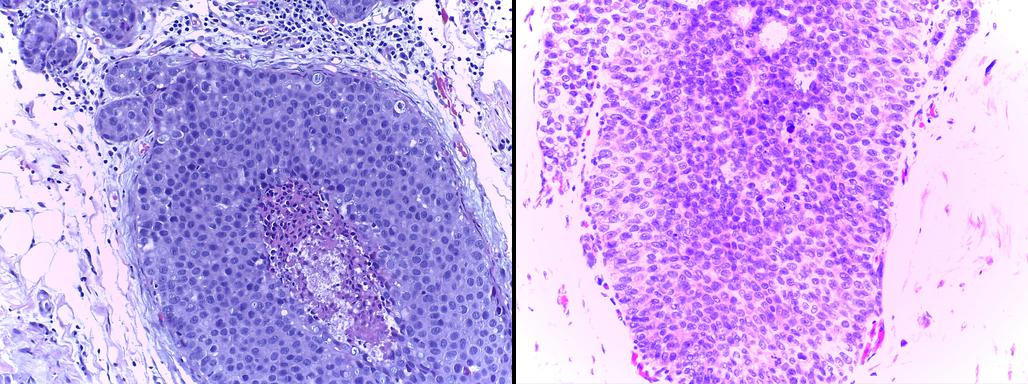}
  \caption{Difference in staining for two images both labeled as \emph{In Situ Carcinoma} extracted from different WSIs~\citep{aresta2018bach}.}
  \label{fig:stain-comparison}
\end{figure}

Histology images present additional challenges for ML/DL models because of their (1) high resolution where, for instance, a single core of prostate biopsy tissue digitized at ${40\times}$ magnification is approximately $(15000 \times 15000)$ elements ($\sim225$ million pixels); (2) heterogeneous nature resulting mainly from the variation of the WSI production process; and (3) noisy/ambiguous labels~\citep{Daisuke201834} caused by the annotation process that is conducted by assigning the worst stage of cancer to the image. Therefore, a WSI that is annotated with a specific grade is also more likely to contain regions with lower grades. This leads to imbalanced datasets having fewer images with high grades. Noisy/ambiguous labels are an issue for models trained through multi-instance learning~\citep{carbonneau2018multiple,CheplyginaBP19MEDIA,wang2018revisiting,zhou2004multi}, where the WSI label is transferred to sampled image patches and can  introduce annotation errors. Such label inconsistencies can degrade model performance, and hinder learning~\citep{frenay2014,sukhbaatar2014training,Zhang2017} (see Figs. \ref{fig:wsisegmentationannot} and \ref{fig:stain-comparison}). 

Training accurate DL models to analyze histology images often requires full supervision to address key tasks, such as classification, localization, and segmentation \citep{Daisuke201834,janowczyk2016deep}. Learning to accurately localize cancerous regions typically requires a large number of images with pixel-wise annotations. Considering the size and complexity of such images, dense annotations come at a considerable cost, and require highly trained experts. Outsourcing this task to standard workers such as Mechanical Turk Workers is not an option. As a result, histology datasets are often composed of large images that are coarsely annotated according to the diagnosis. It is therefore clear that training powerful DL models to simultaneously predict the image class and, localize important image regions linked to a prediction \emph{without} dense annotations; is highly desirable for histology image analysis.
 
Despite their intrinsic challenges~\citep{choe2020evaluating}, techniques for weakly-supervised learning (WSL)~\citep{zhou2017brief} have recently emerged to alleviate the need for dense annotation, particularly for computer vision applications. These techniques are adapted to different forms of weak supervision,  including image-tags (image-levels label)~\citep{kim2017two,pathak2015constrained,teh2016attention,wei2017object}, scribbles~\citep{lin2016scribblesup,ncloss:cvpr18}, points~\citep{Bearman2016s}, bounding boxes~\citep{dai2015boxsup,Khoreva2017}, global image statistics, such as the target size~\citep{bateson2019constrained,jia2017constrained,kervadec2019curriculum,kervadec2019constrained}. The reduced weak supervision requirement provides an appealing learning framework.  In this paper, we focus on WSL methods that allow training a DL model using only image-level annotations for 
the classification of histology images and for the localization of image ROIs linked to class predictions. These methods perform the weakly-supervised object localization (WSOL) task,  which can produce localization under the form of activation maps and bounding boxes~\citep{choe2020evaluating}.

Interpretability frameworks~\citep{Samek2019,ZhangTLT21} have attracted much attention in computer vision~\citep{AlberLSHSMSMDK19,bau2017network,belharbi2020DeepAlJoinClSegWeakAnn,dabkowski2017real,fong2019understanding,Fong2017ICCV,goh2020understanding,Murdoch22071,PetsiukDS18,Petsiuk2020,Ribeiro2016kdd,samek2020toward,zhang2020interpretable}, and medical image analysis~\citep{cruz2013deep,delatorre2020deep,Fan2020,Ghosal2020InterpretableAS,Hagele2020,hao2019page,korbar2017looking,SaleemSR21,TAVOLARA2020103094}. They are related to WSOL in the sense that it also allows providing a spatial map associated with a class prediction decision. However, interpretability methods are often evaluated differently, using for instance the pointing game~\citep{zhang2018top}, which allows localizing an object via a point. Therefore, we limit the focus of this paper to DL models in the literature that were designed and evaluated mainly for the localization task.

Currently, Class Activation Mapping (CAM) methods are practically the only technique for WSOL~\citep{belharbi2022fcam}. CAMs are built on top of convolution responses over an image, leading to a natural emergence of ROIs.  Strong spatial activations in CAMs correspond to  discriminative ROIs~\citep{zhou2016learning} which allow object localization. Note that localization maps in CAM-based methods are part of the model itself. Such methods have been widely studied in the literature for the weakly-supervised object localization task. In parallel, other methods have emerged for the interpretability, explainability, and visualization of machine learning models~\citep{Samek2019}. These methods often provide visualization tools, such as saliency maps to characterize the response of a pre-trained network for the input image. These methods  include approaches such as attribution methods~\citep{dabkowski2017real,Fong2017ICCV,fong2019understanding,PetsiukDS18,zeiler2014}. Different from the CAM, they produce saliency maps that are external to the network architecture, and that are often estimated by solving an optimization problem. In addition to not being commonly used for object localization, these methods have their own evaluation metrics, such as the pointing game~\citep{zhang2018top}.
Apart from CAM methods, ~\citep{MeethalPBG20icprcstn} presents the only work that aims  to directly produce a bounding box, without using any CAMs, in order to localize objects. In~\citep{ZhangCW20rethink}, the authors aim to train a regressor to produce bounding boxes, where the target boxes are estimated from CAMs. Our review has shown that there very few works weakly localize objects without using CAMs, because of the difficulty in  producing a bounding box using only global labels. CAMs have emerged as a natural response of convolution over visible pixels. However, a bounding box is an abstract and invisible shape, making it difficult to produce without explicit supervision, \ie, bounding box target. This difference between the two approaches explains the current state of the literature on WSOL.
Generally, the goal of CAM methods is to build a model that is able to correctly classify an image where only image-class labels are needed. The methods also yield a per-class activation map, \ie, a CAM, under the form of a soft-segmentation map allowing the pixel-wise localization of objects. This  map can also be post-processed to estimate a bounding box. Therefore, the scope of this paper is limited to CAM methods.

In this work, we provide a review of state-of-the-art deep WSOL methods proposed from 2013 to early 2022. Most of these reviewed methods have been proposed and evaluated on natural image datasets, with only few  having been developed and evaluated with histology images in mind. The performance of representative methods is compared using two public histology datasets for breast and colon cancer, allowing to assess their  classification and localization performance. While there have been different reviews of ML/DL models for medical image analysis, and particularly for the analysis of histology WSIs~\citep{Daisuke201834,janowczyk2016deep,kandemir2015computer,LITJENS201760,SUDHARSHAN2019103} and medical video analysis~\citep{quellec2017multiple}, these have focused on fully supervised tasks, semi-supervised tasks, or a mixture of different learning settings for classification and segmentation tasks~\citep{LITJENS201760,srinidhi2019deep}. To our knowledge, this paper represents the first review focused on deep WSOL models, trained on data with image-class labels for the classification of histology images and localization of ROIs.

Deep WSOL methods in the literature are divided into two main categories, based on the flow of information in models, namely, bottom-up and top-down methods. Our review shows that research in bottom-up  methods is more active and dominant than is the case with to top-down methods, making the former state-of-the-art techniques. To address the shortcomings of CAMs, bottom-up methods have progressed from designing simple spatial pooling techniques to performing perturbations and self-attention, to using shallow features, and most recently, to exploiting pseudo-annotation and separating the training of classification from localization tasks. Recent successful WSOL techniques combine the use of shallow features, with pseudo-annotation, while decoupling classification and localization tasks. Top-down techniques for their part have seen less progress. The methods usually rely either on biologically-inspired processes, gradients, or confidence scores to build CAMs.
Our comparative results study revealed that while deep WSOL methods proposed for histology data can yield good results, generic methods initially proposed for natural images nevertheless produced poor results. The former methods often rely on priors that aim to reduce false positives/negatives related to the ROI size, for example, or use explicit pixel-wise guidance collected from pre-trained classifiers. Overall, all WSOL methods suffer from high false positive/negative localization. We discuss several issues related to the application of such methods to histology data, including the under-/over-activation of CAMs, sensitivity to thresholding, and model selection. CAM over-activation is a new behavior that may be caused by the visual similarity between the foreground and the background.
 
In \autoref{sec:survey}, a taxonomy and a review of state-of-the-art deep WSOL methods are provided, followed by our experimental methodology (\autoref{sec:methodology}) and results (\autoref{sec:results}). We conclude this work with a discussion of the main findings, and key challenges facing the application of such WSOL methods in histology, and provide future directions to mitigate these challenges and potentially reduce the gap in performance between WSOL and fully supervised methods. More experimental details are provided in \autoref{sec:hyper-params-search} and \autoref{sec:sampling-camelyon16-tech-details}. In addition, more visual results of localizations are presented in \autoref{sec:visual-results}. Our code is publicly available.

\section{A taxonomy of weakly-supervised object localization methods}
\label{sec:survey}

\begin{figure}[hpt!]
\centering
\includegraphics[width=0.75\linewidth]{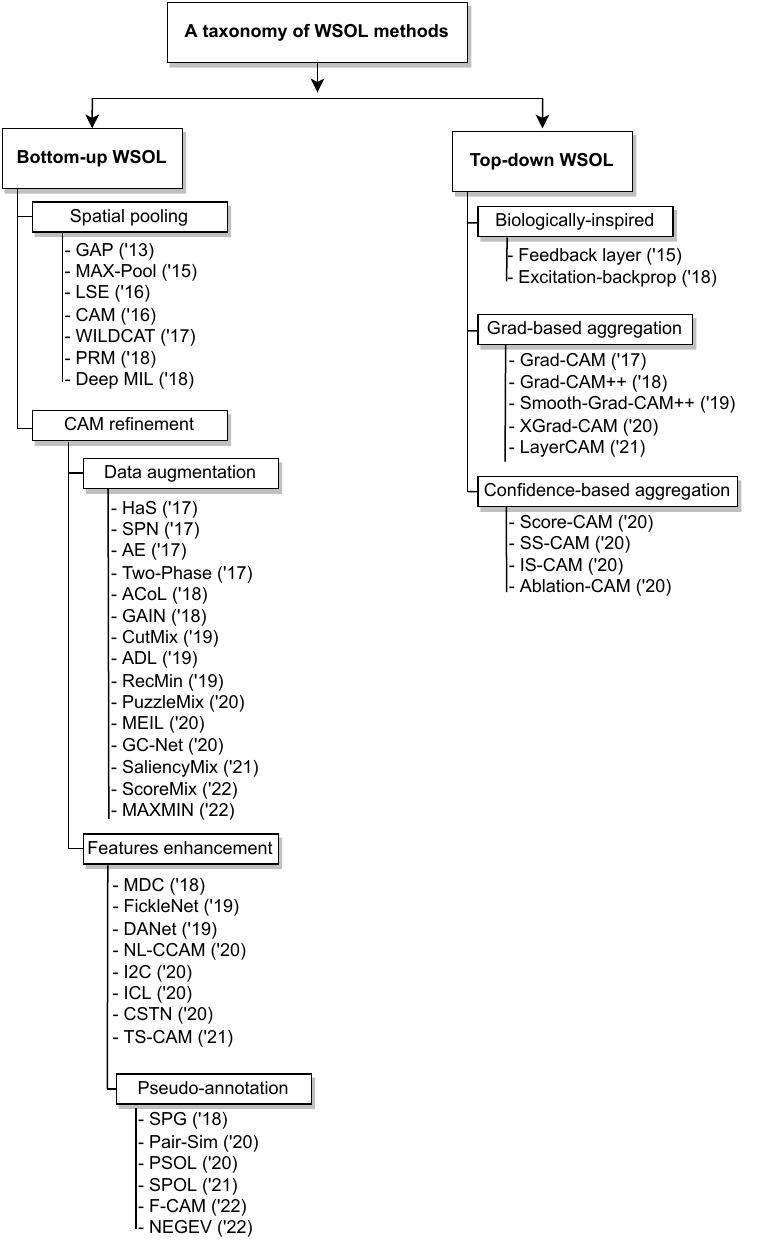}
\caption{Overall taxonomy of deep WSOL methods for training on data with global image-class annotations, and classification and ROI localization. Methods in each category are ordered chronologically: \textbf{1. Bottom-up}: relies on forward pass information. \textbf{2. Top-down}: Exploits both forward and backward pass information.}
\label{fig:fig-taxonomy}
\end{figure}

\begin{figure}[h!]
\centering
\includegraphics[width=.7\linewidth]{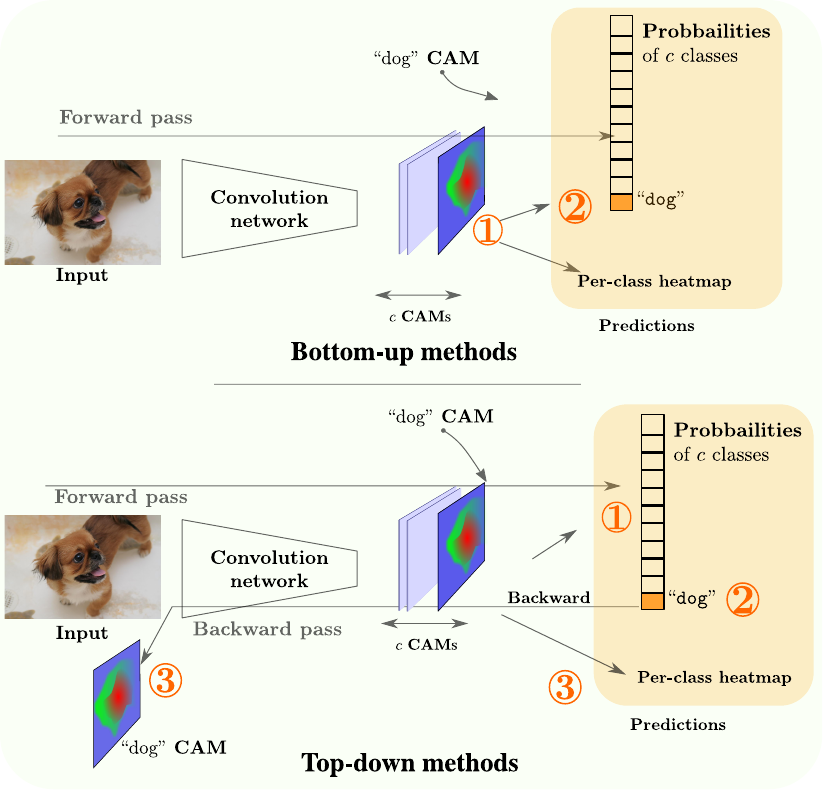}
\caption{Illustration of the main differences between bottom-up (\emph{top}) and top-down (\emph{bottom}) methods for deep WSOL. Both approaches provide CAMs. However, bottom-up techniques produce them during the forward pass, while top-down techniques require a forward, then a backward pass to obtain them. The numbers 1, 2, and 3 in circles indicate the order of operations. The top-down methods can either produce CAMs at the top before the class pooling or the input of the network.
}
\label{fig:fig-diff-flow}
\end{figure}

In our taxonomy, we focus on deep WSL methods that allow classifying an image and \emph{pixel-wise} localize its ROIs via a heat map (soft-segmentation map, CAM)\footnote{In practice, these methods can also yield a bounding box after performing an image processing procedure over the CAM~\citep{choe2020evaluating}.} During training, only global image-class labels are required for supervision. These methods are referred to as weakly-supervised object localization methods (WSOL)~\citep{choe2020evaluating}.

\autoref{fig:fig-taxonomy} illustrates the overall taxonomy. Among deep WSOL methods, we identify two main categories based on the information flow in the network to yield region localization (see \autoref{fig:fig-diff-flow}): 
(a) bottom-up methods, which are based on the forward pass information within a network, and (b) top-down methods, which exploit the backward information in addition to a forward pass.  Each WSOL aims to stimulate the localization of ROI using a different mechanism. Both categories rely on building a spatial attention map that has a high magnitude response over ROI and low activations over the background. The rest of this section provides the notations used herein, details on the main categories and sub-categories, and highlights of the main emerging trends that contributed to the progress of the WSOL task.

\noindent  \textbf{Notation}. 
To describe the mechanisms behind different methods, we introduce the following notation. Let us consider a set of training samples ${\mathbb{D}=\{(\bm{x}^{(t)}, y^{(t)})\}}$ of images ${\bm{x}^{(t)} \in \R^{D\times H^\text{in}\times W^\text{in}}}$ with $H^\text{in}$, $W^\text{in}$, $D$ being the height, width and depth of the input image, respectively; its image-level label (\ie, class) is $y^{(t)} \in \Y$,  with $C$ possible classes. For simplicity, we refer to a training sample (an input and its label) as ${(\bm{x}, y)}$.

Let ${f_{\bm{\theta}}:\R^{D\times H^\text{in}\times W^\text{in}} \to \Y}$ be a function that models a neural network, where the input ${\bm{x}}$ has an arbitrary height and width and ${\bm{\theta}}$ is the set of model parameters. The training procedure aims to optimize parameters ${\bm{\theta}}$ to achieve a specific task. In a multi-class scenario, the network typically outputs a vector of scores ${\bm{s} \in \R^C}$ in response to an input image. This vector is then normalized to obtain a posterior probability using a $\mathrm{softmax}$ function,
\begin{equation}
\label{eq:eq-0000}
\Prob(y = i| \bm{x}) = \softmax(\bm{s})_i = \frac{\exp(\bm{s}_i)}{\sum_{j=1}^C \exp(\bm{s}_j)} \; .
\end{equation}
The model predicts the class label corresponding to the maximum probability: $\argmax \{ \Prob(y=i|\bm{x}): i = 1, 2, ..., C \} = \argmax \{ \bm{s}_i: i = 1, 2, ..., C \}$.

Besides the classification of the input image, we are also interested in the pixel-wise localization of ROIs within the image. Typically, a WSOL method can predict a set of $C$ activation maps of height $H$ and width $W$ to indicate the location of the regions of each class. We note this set as a tensor of shape $\bm{M} \in \R^{C\times H\times W}$, where ${\bm{M}_c}$ indicates the $c^{\text{th}}$ map. ${\bm{M}}$ is commonly referred to as \emph{Class Activation Maps} (CAMs). Due to convolutional and downsampling operations, typical CAMs have a low resolution as compared to the input image. We note the downscale factor as $S$, such that $H = H^\text{in} / S$ and $W = W^\text{in} / S$. Interpolation is often required to yield a CAM of the same size as the image.

\subsection{Bottom-up WSOL techniques}

 In bottom-up methods, the pixel-wise localization is based on the activation of the feature maps resulting from the standard flow of information within a network from the input layer into the output layer (forward pass, \autoref{fig:fig-diff-flow} (\emph{top})). Within this category, we identify two different subcategories of techniques to address weakly supervised localization. The first category contains techniques that rely mainly on spatial pooling. Different ways were proposed to pool class scores while simultaneously stimulating a spatial response in CAM to localize ROIs. These methods had limited success. Therefore, another type of method emerged and aimed to refine CAMs directly while using simple spatial pooling techniques. In the next subsections, we present these methods and their variants.

\subsection{Spatial pooling methods}
\label{subsec:spatial-pooling}

This family of techniques aims to design different spatial pooling methods to compute per-class scores, which are then used to train the whole network for classification using standard cross-entropy. In some cases, the pooling is performed to build an image representation, \ie, bag features. Such spatial pooling allows building maps (CAMs) to localize ROIs. Each method promotes the emergence of ROIs localization differently. This strategy undergirds WSOL, and is considered a pioneering mechanism that introduced weakly supervised localization in deep models~\citep{lin2013network}. Learning preserving spatial information in CAMs allows ROIs \emph{localization} while requiring only global class annotation. Different methods have been proposed to compute the class scores from spatial maps, whith each pooling strategy having a direct impact on the emerging localization. The challenge is to stimulate the emergence of just ROI in the CAM.
All techniques usually start off the same way: a CNN extracts $K$ feature maps $\bm{F} \in \R^{K\times H\times W}$, where $K$ is the number of feature maps, which is architecture-dependent. The feature maps $\bm{F}$ are then used to compute a per-class score using a spatial pooling technique.

The first method is Global Average Pooling (GAP)~\citep{lin2013network}. It simply averages each feature map in ${\bm{F} \in \R^{K\times H\times W}}$ to yield the per-map score in order to build the global representation ${\bm{f} \in \R^K}$ of the input image,
\begin{equation}
\label{eq:eq000009}
\bm{f}_k = \frac{1}{H\,W}\sum\limits_{i=1, j=1}^{H, W} \bm{F}_{k,i,j} \; ,
\end{equation}
where $\bm{f}_k$ is the $k^{\text{th}}$ feature of the output.
The class-specific activations are then obtained by a linear combination of the features using the weights of the classification layer. Note that in practice, one can directly average CAMs, when available, to yield per-class scores instead of using an intermediate dense layer. In both cases, this pooling strategy ties the per-class score to \emph{all} spatial locations on a map. This means that both ROIs and the background participate in the computation of the per-class score.
The CAM literature shows that this pooling strategy can be used to allow a CNN to perform localization using only global labels~\citep{zhou2016learning}. Typically, in a CNN, the last layer which classifies the representation $\bm{f}$ is a fully connected layer parameterized by $\bm{W}\in\R^{C\times K}$ such that $\bm{s} = \bm{W}\bm{f}$ (bias is omitted for simplicity). The CAMs, denoted as ${\bm{M}\in\mathbb{R}^{C\times H\times W}}$, are then obtained using a weighted sum of the spatial feature ${\bm{F}}$,
\begin{equation}
\label{eq:cams}
\bm{M}_c = \sum\limits_{k=1}^{K}\bm{W}_{c,k} \, \bm{F}_k \; .
\end{equation}
This strategy has been widely used for natural scene images as well as for medical images \citep{feng2017discriminative,gondal2017weakly,izadyyazdanabadi2018weakly,sedai2018deep}.

An early work on CAM methods~\citep{zhou2016learning} revealed a fundamental issue, namely, under-activation. CAMs tend to activate only on small discriminative regions, and therefore, localizing only a small part of the object while missing a large part of it. This leads to high false negatives. Subsequent works in WSOL aimed mainly to tackle this issue by pushing activations in CAM to cover the entire object. This is done either through a different pooling strategy or by explicitly designing a method aiming to recover the full object (\autoref{subsec:cam-refinement}).

As an alternative to averaging spatial responses, authors in~\citep{oquab2015object} consider using the \emph{maximum} response value on the CAM as a per-class score (MAX-Pool). The method therefore avoids including potential background regions in the class score, thus reducing false positives. However, this pooling technique tends also to focus on small discriminative parts of objects since the per-class score is tied only to one pixel of the response map\footnote{Note that one pixel in the CAM corresponds to a large surface in the input image depending on the size of the receptive field of the network at the CAM layer.}. To alleviate this problem ~\citep{pinheiro2015image,sun2016pronet} consider using a smoothed approximation of the maximum function to discover larger parts of objects of interest using the \emph{log-sum-exp} function (LSE), 
\begin{equation}
\label{eq:eq0002}
 \bm{s}_c = \frac{1}{q}\log\big[ \frac{1}{H\,W} \sum_{i=1,j=1}^{H,W} \exp(q\, \bm{M}_{c,i,j})\big] \; ,
\end{equation}
where $q\in\R_+^*$ controls the smoothness of the approximation. A small $q$ value makes the approximation closer to the average function, while a large $q$ makes it close to the maximum function. Thus, with small  $q$  values, make the network consider large regions, while large values consider only small regions.
        
Instead of considering the maximum of the map~\citep{oquab2015object}, \ie, a single high response point, authors in~\citep{ZhouZYQJ18PRM} (PRM) propose to use \emph{local maxima}. This amounts to using local peak responses which are more likely to cover a larger part of the object than occurs when using only the single maximum response,
\begin{equation}
\label{eq:prm-score}
\bm{s}_c = \bm{M}^c * G^c = \frac{1}{N^c}\sum^{N^c}_{k=1}\bm{M}^c_{i_k,j_k}\; ,
\end{equation}
where $G^c \in \mathbb R^{H \times W}$ is a sampling kernel, $*$ is the convolution operation, and $N^c$ is the number of local maxima. Depending on the size of the kernel, this pooling allows stimulating different distant locations, which can help recover adjacent regions with an object. Similarly, it is likely that background regions are stimulated. This highlights the challenge faced with transferring global labels into local pixels. Note that such a transfer of supervision is known to be an ill-posed problem in the field of WSOL~\citep{wan2018min}.

All the pooling methods discussed thus far rely on high responses to yield per-class scores. The assumption with CAMs is that strong responses indicate potential ROIs, while low responses are more likely to represent  backgrounds. This assumption is incorporated in the computation of per-class scores, and therefore, has a direct impact on the localization in CAMs. Authors in~\citep{durand2017wildcat,durand2016weldon} (WILDCAT) pursue a different strategy by including low activation, \ie, negative evidence, in the computation of the per-class score. They argue that such pooling plays a regularization role and prevents overfitting, allowing better classification performance. However, it remains unclear how tying negative regions to class scores improves localization, since the aim is to maximize the per-class score of the true label. Nevertheless, the authors provide an architectural change of the pooling layer where several \emph{modality maps} per class are considered. Hence, these modalities allow to capture several parts of the object, leading to better localization. Formally, the pooling is written as, 
\begin{equation}
\label{eq:wildcat-score}
\bm{s}_c = \frac{Z_c^+}{n^+} + \alpha \frac{Z_c^-}{n^-} \; ,
\end{equation}
where $Z_c^+$ and $Z_c^-$ correspond to the sum of the $n^+$ highest and $n^-$ lowest activations of $\bm{M}_c$ respectively, and $\alpha$ is a hyper-parameter that controls the importance of the minimum scoring regions. Such an operation consists in selecting for each class the ${n^+}$ highest activation and the ${n^-}$ lowest activation within the corresponding map.
This method has also been used in the medical field for weakly supervised region localization and image classification in histology images~\citep{belharbi2019weakly,belharbi2022minmaxuncer}.  In~\citep{courtiol2018classification}, instead of operating on pixels, the authors consider adapting~\citep{durand2017wildcat,durand2016weldon} for WSIs to operate on instances (tiles). 
    
The aforementioned methods build a bag (image) representation, and then compute CAMs that hold local localization responses, and finally, pull the per-class scores. Authors in~\citep{ilse2018attention} (Deep MIL) rely explicitly on a multi-instance learning (MIL) framework~\citep{carbonneau2018multiple, CheplyginaBP19MEDIA, wang2018revisiting, zhou2004multi}. Here, instance representations are firstly built. Then, using the attention mechanism~\citep{BahdanauCB14corrattention}, a bag representation is computed using a weighted average of the instances representations. In this case, it is the attention weights that represent the CAM. Strong weights indicate instances with ROIs, while small weights indicate background instances. This method requires changes to standard CNN models. In addition, it is tied to binary classification only. Adjusting to a multi-class context requires further changes to the architecture. Formally, given a set of features $\bm{F}\in\R^{K\times H\times W}$ extracted for an image, the representation $\bm{f}$ of the image is computed as,
\begin{align}
  &\bm{f} = \sum\limits_{i=1,j=1}^{H,W}\bm{A}_{i,j}\bm{F}_{i,j} \; ,\label{eq:eq0001} \\
  \text{and} \quad & \bm{A}_{i,j} = \frac{\exp(\psi(\bm{F}_{i,j}))}{\sum_{i=1,j=1}^{H,W}  \exp(\psi(\bm{F}_{i,j}))} \; , \nonumber
\end{align}
where $\bm{F}_{i,j}$ is the feature vector of the location (\ie, instance) indexed by $i$ and $j$. ${\psi:\R^K\to\R}$ is a scoring function. The resulting representation $\bm{f}$ is then classified by a fully connected layer. Two scoring functions are considered~\citep{ilse2018attention},
\begin{align}
  \psi_1(\bm{f}) &= \bm{w}\tanh(\bm{V}\bm{f}) \; ,\\
  \psi_2(\bm{f}) &= \bm{w}\big[\tanh(\bm{V}\bm{f})\odot\sigma(\bm{U}\bm{f})\big] \; ,
\end{align}
where $\bm{w}\in\R^{L}$, $(\bm{V}, \bm{U})\in\R^{L\times K}$ are learnable weights, and ${\odot}$ is an element-wise multiplication.
This approach is designed specifically for binary classification and produces a matrix of attention weights ${\bm{A}\in[0, 1]^{H\times W}}$ with ${\sum\bm{A}=1}$. In the next section, we present a second bottom-up category that aims to refine the CAMs directly.

 \subsubsection{CAM refinement methods} 
 \label{subsec:cam-refinement}
 
While spatial pooling methods have helped the emergence of some discriminative regions in CAMs, they have limited success when it comes to covering the full foreground object. Under-activation of CAMs is still a major ongoing issue in the WSOL field, reflecting the difficulty face in to transferring global labels to pixel level. Ever since this became clear, research has shifted from improving the pooling function to explicitly overcoming the under-activation issue and recovering the entire object. Often, this is achieved while using simple pooling functions such as the GAP method~\citep{lin2013network}, and to this end, different strategies have been proposed. We divide these into two main categories: methods that use data augmentation to \emph{mine} more discriminative regions, and methods that aim to enhance and learn better internal features of a CNN.
 
\noindent \textbf{Data augmentation methods.} Data augmentation is a strategy often used in machine learning to prevent overfitting and improve performance~\citep{Goodfellow-et-al-2016}. It has been similarly been used in the WSOL field to prevent models from overfitting one single discriminative region, \ie, from under-activating. This is often achieved by pushing the model to seek, \ie, \emph{mine}, other discriminative regions, thereby promoting a large coverage of objects in CAM. Data augmentation most commonly takes the form as information suppression, \ie, \emph{erasing}, where part of an input signal is deleted. This can be performed over input images or intermediate features. Conceptually, this can be seen as a \emph{perturbation} process to stimulate the emergence of more ROIs. 
For instance, authors in~\citep{SinghL17} propose a 'Hide-And-Seek' (HaS) training strategy, where the input image is divided into multiple patches. During the training phase, only some of these patches are randomly set to be visible while the rest are hidden. Such data augmentation has already been shown to regularize CNN models and improve their classification performance~\citep{devries2017cutout} (cutout). This is similar to applying a dropout~\citep{srivastava14dropout} over the input image, where the target regions consist of a whole patch instead of a single pixel. As a result, the network will not overly rely on the most discriminative patches, and will seek other discriminative regions. While this is an advantage, it can be counter-productive as the network may inadvertently be pushed to consider the background as discriminative, especially for small objects that can be easily deleted.

Other data augmentations have been exploited to improve localization. For instance, the MixUp method~\citep{ZhangCDL18mixup} was designed to regularize neural networks by making them less sensitive to adversarial examples and reducing their memorization. This is done by blending two training images to certain degree, in which case the label of the augmented image is assigned by the linear combination of the labels of the two images. Despite the augmented images looking unnatural and locally
ambiguous, the method improves classification accuracy.  Authors in~\citep{YunHCOYC19} (CutMix) adapt this method to improve localization. Instead of fully blending images, they propose to randomly cut a patch from an image and mix it with a second image. The label is mixed proportionally to the size of the patch. In essence, this is similar to cutout~\citep{devries2017cutout} and HaS~\citep{SinghL17}, but instead patches being filled with black or random noise, they are filled with pixels from another image. In practice, this has been shown to improve localization performance. However, due to the randomness in the source patch selection process, this method may select background regions leading to wrong mixed labels, which then leads to the classifier learning unexpected feature representations.
Similarly, \citep{KimCS20puzzlemix} proposed PuzzleMix, which jointly optimizes two objectives: selecting an optimal mask and selecting an optimal mixing plan. Here, the mixing of the input images is no longer random, but uses image saliency, which emerges from image statistics. The mask tries to reveal the most salient data of the two images. Meanwhile, the optimal transport plan aims to maximize the saliency of the revealed portion of the data. In the same vein, SaliencyMix~\citep{UddinMSCB21} exploits image saliency, but uses a bounding box to capture a region remix instead of a mask. Note that relying on image saliency is a major drawback for less salient images such as those bearing histology data since the foreground and background look similar. Authors in~\citep{stegmuller2022} (ScoreMix) applied this type of approach to histology data by using proposed regions via attention. Mixing region approach is based on classifier attention instead of image statistics. Discriminative regions from the sources are cut and mixed over non-discriminative regions of the target. Conceptually, this gives a better regional mixing. However, since the learned attention can easily hold false positives/negatives, the mixing can still be vulnerable. In addition, the obtained results seem relatively close to those of the CutMix method~\citep{YunHCOYC19}.

In~\citep{wei2017object} (AE), the authors propose an iterative strategy to mine discriminative regions for semantic segmentation. Similarly to the HaS method~\citep{SinghL17}, they erase regions with the highest response values through learning epochs of a classifier. This allows the emergence of large parts of the model. The emerging segmentation proposals are used to train the model for semantic segmentation. Sequential erasing yields a computationally expensive process since multiple rounds are required. To improve this, ACoL~\citep{ZhangWF0H18} designed two branch classifiers to predict the discriminative region and corresponding complementary area simultaneously. The MEIL method~\citep{MaiYL20eil} proceeds in a similar fashion by adding multiple output branches that exploit the erasing process within the learning. 

Guided Attention Inference Network (GAIN)~\citep{LiWPE018CVPR} method uses two sequential networks with a shared backbone to mine ROIs. The first network yields an attention map of ROIs, which is used to erase discriminative regions in the image. The erased image is then fed into the next network, where its class response with respect to the target label is used to ensure that no discriminative regions are left in the image after the erasing process. The ROI suppression process is expected to push the first model to seek more discriminative regions, hence large ROIs are covered by the CAM. Similarly, authors in~\citep{kim2017two} (Two-Phase) consider two-phase training of two networks. The first network is trained until convergence. Then, it is used, with frozen weights, in front of a second network to produce a CAM of the target label. The CAM is thresholded to localize the most discriminative regions. Instead of masking the input image as done in the GAIN method~\citep{LiWPE018CVPR}, the authors consider masking intermediate feature maps. Once again, results show that this type of information hiding at the feature level allows exploring more ROIs to uncover complete objects.

The GC-Net method~\citep{LuJXSZ020gcnet} considers incorporating Geometry Constraints (GC) to train a network to localize objects. Specifically, the authors use 3 models: a detector that yields object localization under the form of a box or an ellipse; a mask generator, which generates a mask based on the generated localization, and a classifier that is evaluated over the ROIs covered by the mask and its complement, \ie, background. The detector is trained to produce small ROIs in which the classifier has a high score while a low score is achieved over the background.

Authors in~\citep{belharbi2019weakly} (RecMin) consider a recursive mining algorithm integrated directly into back-propagation, allowing to mine ROIs  on the fly. All these methods perform mining-erasing of information over the input image. The ADL~\citep{ChoeS19} method builds a self-attention map per layer to spot potential ROIs. Then, it stochastically erases locations over multiple intermediate feature maps at once during forward propagation through simple element-wise multiplication. The erasing is performed by simple dropout over the attention mask. Such a procedure allows the enhancement of both classification and localization performance. Note that self-attention was already used prior to ADL in~\citep{zhu2017soft} (SPN) as a layer to yield proposal regions that are coupled with feature maps allowing only potential ROIs to pass to the next layer, filtering out background/noise.
Authors in~\citep{belharbi2022minmaxuncer} (MAXMIN) use two models: a localizer, followed by a classifier. The localizer aims to build a CAM to localize ROIs at the pixel level. The input image is masked by the produced CAM, and then fed to the classifier. The authors explicitly include the background prior to learning the CAM by constraining it to holding both foreground and background regions. This prevents under-/over-activations, which in turn reduces false positives/negatives. Using entropy, the target classifier scores are constrained to be low over the background and high over the foreground, thus ensuring that no ROI is left in the background. 
A significant downside of these erasing/mining-based methods is their inherent risk of over-mining since there are no clear criteria to stop mining.  While they are efficient at expanding and gathering object regions, it is very easy to expand to non-discriminative regions, which directly increases false positives.

\noindent \textbf{Features enhancement methods.} Other methods aim to improve localization by learning better features. This is often achieved through architectural changes of standard models or by exploiting different levels of features for localization, such as shallow features. Additionally, using pseudo-labels to explicitly guide learning has emerged as an alternative approach for tackling WSOL tasks.

Authors in~\citep{WeiXSJFH18} analyze the impact of the object scale on predictions and propose to exploit dilated-convolution~\citep{ChenPKMY18,ChenPKMY14}. They equip a classifier with a varying dilation rate: multi-dilated convolutional (MDC) blocks. This has been shown to effectively enlarge the receptive fields of convolutional kernels, and more importantly, to transfer the surrounding discriminative information to non-discriminative object regions, promoting the emergence of ROI while suppressing the background.
Unlike most works that pull CAMs from the top layer (high level), authors in~\citep{YangKKK20} (NL-CCAM) consider non-local features by combining low- and high-level features to promote better localization. In addition, rather than using a per-class map as the final CAM, they combine all CAMs using a weighted sum after ordering them using their posterior class probabilities. This allows to gather several parts of the objects and to suppress background regions in the final localization map.
The FickleNet method~\citep{LeeKLLY19} randomly selects hidden locations over feature maps. During training, for each random selection, a new CAM is generated.  Therefore, for each input image, multiple CAMs can be generated to predict the most discriminative parts. This allows building CAMs that better cover the object. This method is related to ADL~\citep{ChoeS19}, which uses attention, followed by dropout, to mask features. FickleNet does not rely on attention, and simply drops random locations.

DANet~\citep{XueLWJJY19iccvdanet} uses multi-branch outputs at different layers to yield a CAM with different resolutions. This allows to obtain a hierarchical localization. To spread activation over the entire object \emph{without} deteriorating the classification performance, the authors consider a joint optimization of two different terms. A discrepant divergent activation loss constrains CAMs of the same class to cover \emph{different} regions. The authors note that classes with similar visual features are typically suppressed in standard CNNs, since the latter are not discriminative. To recover these regions, they propose a hierarchical divergent activation loss. Meta-classes are created hierarchically to gather previous meta-classes, in which the bottom of the hierarchy contains the original classes. At a specific level, the classifier is trained to assign the same meta-class for all samples assigned to it. This pushes shared similar features to activate within that meta-class, hence recovering similar features in original classes.

In the I${^2}$C method~\citep{ZhangW020i2c}, the authors propose to leverage pixel-wise similarities at the spatial feature level via Inter-Image Communication (I${^2}$C) for better localization. Local and global discriminative features are pushed to be consistent. A local constraint aims to learn the stochastic feature consistency among discriminative pixels, which are randomly sampled from a pair of images within a batch. A global constraint is employed, where a global center feature per-class is maintained and updated in memory after each mini-batch. Average local random features are constrained to be close to the center class features.
The ICL method~\citep{KiU0B20icl} aims to deal with over-activation by preventing CAMs from spiking over the background. An attention contrastive loss is proposed. Similar to ADL~\citep{ChoeS19}, an attention map is estimated from feature maps. Very high and very low activations are used to estimate potential foreground and background regions. The middle activations could be either foreground or background. To expand the activation from foreground into uncertain region, the contrastive loss aims to push activation with \emph{spatial features} similar to foreground features to be foreground while activations with similar spatial features to background features are pushed to be background. This allows a careful expansion of foreground activation toward background regions. In addition, attention at the top layer, which is semantically rich, is used in a self-learning setup to align and guide low layer attention, which is often noisy.

The WSOL task has also benefited from recent advances in architectural design in deep learning. Transformers~\citep{DosovitskiyB0WZ21} in particular have seen their first use in such a task in the  TS-CAM~\citep{gao2021tscam} method. A visual transformer~\citep{DosovitskiyB0WZ21} constructs a sequence of tokens by splitting an input image into patches with positional embedding and applying cascaded transformer blocks to extract a visual representation. Visual transformers can learn complex spatial transforms and reflect long-range semantic correlations adaptively via self-attention mechanism and multilayer perceptrons. This occurs to be crucial for localizing full object. TS-CAM~\citep{gao2021tscam} improves patch embeddings by exploiting self-attention. In addition, a class-agnostic map is built at each layer. The authors equipped the transformers' output with a CAM module allowing to obtain semantic maps. The Final CAM is an aggregation between the CAM yielded by the CAM module and the average class-agnostic maps across all layers. This shows to help improve localization.
In the CSTN method~\citep{MeethalPBG20icprcstn}, the authors replace standard convolution filters with Spatial Transformer Networks (STNs)~\citep{jaderberg2015spatial}. In addition to using multi-scale localization, this STN model learns affine transformations, which can cover different variations including translation, scale, and rotation, allowing to better attend different object variations.

Recently, a new trend has emerged in WSOL, in which \emph{pseudo-annotations} are exploited. An external model or a WSOL classifier is initially trained  using weak labels. Then, it is used to collect pseudo-labels which represent a substitute for the missing full supervision. They are then used to fine-tune a final model. This provides explicit localization guidance for training. However, such methods inherit a major drawback in the form of learning with inaccurate/noisy labels, which must be dealt with.
For instance, in \citep{RahimiSAHB20} (Pair-Sim), the authors use a fully supervised source dataset to train a proposal generator (Faster-RCNN~\citep{RenHGS15}). Then, they apply the generator over a target weakly supervised dataset for the WSOL task to yield proposals for each sample, \ie, bag. The classical MIL framework~\citep{carbonneau2018multiple, CheplyginaBP19MEDIA, wang2018revisiting, zhou2004multi} is applied by splitting the target dataset into 2 subsets conditioned on the class; one with positive samples with that class label, and the other holding negative samples. The MIL framework is solved such as to yield exactly one proposal per positive sample. A unary score regarding the proposal abjectness that is learned from the source is used, in addition to a pairwise score that measures the compatibility of two proposals conditioned on the bag class.

Authors in~\citep{ZhangCW20rethink} show that localization and classification interfere with each other in WSOL, and that these should be divided into two separate tasks. They first train a classifier, which is used to generate Pseudo Supervised Object Localization (PSOL). This pseudo-supervision is then used to train a separate class-agnostic localizer. In the same vein, the work in~\citep{wei2021shallowspol} demonstrates the benefits of shallow features for localization. The authors exploit low level (Shallow) features to yield Pseudo supervised Object Localization (SPOL), which is used to guide the training of another network. The F-CAM method~\citep{belharbi2022fcam} also exploits shallow features by equipping standard classifiers with a segmentation decoder to form a U-Net architecture~\citep{Ronneberger-unet-2015}. Such a model builds the final CAM through top and low features using skip-connections. 
The authors show the impact of the CAM size on localization performance, with a lower localization performance seen with a smaller CAM size. CAMs are often interpolated to have the same size as the input image. Since the interpolation algorithm does not take into consideration the image statistics, the authors propose to gradually increase the resolution of the CAMs via a parametric decoder. A low resolution CAM, image statistics and generic size priors are used to train the decoder. The authors propose a \emph{stochastic} sampling of local evidence as opposed to common practice in the literature, where pseudo-labels are selected and fixed before training. The F-CAM method was further adapted for transformer-based methods~\citep{murtaza2022wacvw,murtaza2022mtlai} for WASOL in drone-surveillance, and subsequently, for WSOL in videos~\citep{tcamsbelharbi2023}.
Following F-CAM architecture, NEGEV~\citep{negevsbelharbi2022} was proposed for histology data to improve localization and classifier interpretability. However, the authors focus mainly on using negative evidence collected from a pre-trained classifier, as well as evidence occurring naturally in datasets, \ie, fully negative samples. This allows the method to achieve state-of-the-art performance in localization. Additional experiments also show  that the stochastic sampling proposed in~\citep{belharbi2022fcam} outperforms the fixed selection of local evidence by a large margin.
The Self-Produced Guidance method (SPG)~\citep{ZhangWKYH18} extracts several attention maps from the top- and low-level layers to benefit from global and detailed localization. Discrete locations of potential ROIs are collected from the maps using thresholding, and are then used to train different layers in a self-supervised way. Note that this approach is related to SPN~\citep{zhu2017soft} and ADL~\citep{ChoeS19}, which   exploit attention as a self-guidance mechanism to steer the focus toward potential ROIs by masking feature maps. However, the SPG method explicitly learns a segmentation mask using discrete pixel-wise information collected from attention as supervision.

\subsection{Top-down WSOL techniques}
  
This second main category is based essentially on the backward pass information within a network to build an attention map that localizes ROIs with respect to a selected target class (backward pass, \autoref{fig:fig-diff-flow} (\emph{bottom})). We distinguish three main sub-categories which differ in the way the top signal is back-traced. The first category exploits a secondary conductive feedback network; the second relies on gradient information to aggregate backbone spatial feature maps, while the last exploits posterior class scores for aggregation.

\noindent \textbf{Biologically-inspired methods.} These methods are often inspired by cognitive science. For instance, authors in~\citep{cao2015look} argue that visual attention in humans is typically dominated by a target, \ie, a 'goal', in a top-down fashion. Biased competition theory~\citep{Beck2009,Desimone1998,Desimone95} explains that the human visual cortex is enhanced by top-down stimuli and irrelevant neurons are suppressed in \emph{feedback loops} when searching for objects. The work in~\citep{cao2015look} mimics this top-down loop with a \emph{feedback network} that is attached to standard feedforward networks and holds binary variables in addition to ReLU activations~\citep{NairH10}. These binary variables are activated by a top-down message. Given a target class, a standard forward step is performed within the feedforward network to maximize the posterior score of the target. Then, a backward pass is achieved within the feedback network. To promote \emph{selectivity} in the feedback loop~\citep{Desimone95}, the L${_1}$ norm is used as a sparsity term over the binary variables of the feedback network. The forward/backward process is performed several times in order to optimize a loss function composed of the posterior score of a target class, and the sparsity term over the binary variables. For a localization task, the backward loop can reach the network input layer to yield an attention map that indicates ROIs associated with the target class selected at the top of the network. Despite the benefits of this method for CNN visualization and ROI localization, its iterative optimization process to obtain localization makes it less practical for WSOL tasks.
The excitation-backprop~\citep{zhang2018top} method follows a similar top-down scheme. In particular, the authors consider a top-down Winner-Take-ALL (WTA) process~\citep{Tsotsos95} which selects one winning path. To avoid selecting one deterministic path, which is less representative and leads to binary maps, the authors propose a \emph{probabilistic} WTA downstream process that models all paths. This process integrates both bottom-up and top-down information to compute the winning probability for each neuron. To improve the localization accuracy of the attention map, particularly for images with multi-objects, the authors propose a contrastive top-down attention which captures the differential effect between a pair of top-down attention signals. This allows the attention map to hold activations only for one target class.
While both methods~\citep{cao2015look,zhang2018top} yield good results, they require substantial changes to standard CNN architectures. In addition, the methods are often used for interpretability and explainability of deep models~\citep{Samek2019}.

In the next two categories, we describe how intermediate spatial feature maps are used to pull discriminative maps to localize ROIs associated with a fixed target class. Usually, an aggregation scheme via a weight-per-feature map is performed. The key element in these methods is how the weighting coefficients are estimated. All these weights are back-streamed from the per-class posterior score at the top of the network.
  
\noindent \textbf{Grad-based aggregation.} This family of methods relies on the gradient of posterior class scores of the true label with respect to the feature maps to determine aggregation weights. Such approaches are also used as \emph{visual tools} to explain a network's decision\citep{Samek2019}. In~\citep{SelvarajuCDVPB17iccvgradcam}, the authors propose the Grad-CAM method. In order to compute the CAMs, they propose to \emph{aggregate} spatial feature maps using gradient coefficients. The coefficient of each feature map is computed using the gradient of the score of the selected target class with respect to that map. This gradient indicates how much a pixel location contributes to the target output. The CAM for the class $c$ is computed as,
  \begin{align}
    \bm{M}_c &= \relu\Big(\sum\limits_{k=1}^K \bm{A}_{c,k} \; \bm{F}_k\Big) \;, \\
    \text{where} \quad \bm{A}_{c,k} & = \frac{1}{H\,W} \sum\limits_{i=1,j=1}^{H,W} \frac{\partial \bm{s}_c}{\partial \bm{F}_{k,i,j}} \; ,
  \end{align}
 where ${\bm{s}_c}$ is the score for the class $c$. This approach is a generalization of the CAM method~\citep{zhou2016learning}, where the derivative of the score with respect to the feature map is used instead of learned weights. This approach was improved in Grad-CAM++~\citep{ChattopadhyaySH18wacvgradcampp} and Smooth-Grad-CAM++~\citep{omeiza2019corr} to obtain better localization by covering a complete object, as well as explaining the occurrence of multiple instances in a single image. In \citep{fu2020axiom}, the authors propose theoretical grounds for CAM-based methods, in particular Grad-CAM, for a more accurate visualization. They include two important axioms: sensitivity and conservation, which determine how to better compute the importance weight of each feature map. Following these axioms, the authors propose a new gradient-based method, XGrad-CAM, which computes the coefficient differently. The coefficients are the solution to an optimization problem composed of the two axioms.
 To date, these methods have aggregated the feature maps at the last layer. LayerCAM method~\citep{JiangZHCW21layercam} exploits top- and low-level layers to extract localization. Top layers are commonly known to hold coarse localization, while low-level layers hold detailed but noisy localization. This method extracts CAMs from each layer using a back-propagated gradient signal. Then, the final CAM is computed by fusing all CAMs estimated at each layer.
 Note that this family of methods is designed to \emph{interrogate} a pre-trained model. While they are model-independent, and allow inspecting the decision of a trained model, the user cannot control their behavior during the training of the model. This ties the localization performance of these methods strongly  to the localization information provided by the trained model.

\noindent \textbf{Confidence-based aggregation.} Several methods exploit raw scores of the target class, instead of the gradient, in order to \emph{aggregate} spatial features of a backbone. In the Score-CAM method~\citep{WangWDYZDMH20scorecam}, each feature map is element-wise multiplied with the input image, and is then used to compute the target-class score. This allows one to obtain a posterior score for each spatial map, which is then compared to the score of the raw image. The difference between both scores yields Channel-wise Increase of Confidence (CIC), in which where high values indicate the presence of a strong ROI in the feature map. The final CAM is a linear weighted sum of all the feature maps, followed by ReLU non-linearity~\citep{NairH10} where CIC coefficients are used as weights,
\begin{equation}
    \bm{M}^c_{Score-CAM} = ReLU\Big(\sum_k \alpha^c_k\bm{F}_k^l\Big)\;, 
\end{equation}
where ${\alpha^c_k}$ is the CIC of class $c$ for feature map $k$, and $l$ is a layer. Smoothed-Score-CAM (SS-CAM)~\citep{naidu2020sscam} improves the Score-CAM~\citep{WangWDYZDMH20scorecam}.  Instead of computing the CIC over a single masked image, SS-CAM averages many perturbed masked images. The authors propose either to perturbate the feature map or the input image. This yields smoothed aggregation weights. The IS-CAM method~\citep{naidu2020iscam} performs a similar process to smooth weights. While these methods yield good localization, a recent empirical evaluation showed that they are computationally expensive~\citep{belharbi2022fcam}. For instance, computing a single CAM for an image of size ${224\times224}$ over the ResNet50~\citep{heZRS16} model takes a few minutes on a decent GPU. This makes training such methods impractical.
In the Ablation-CAM method~\citep{desai2020ablation}, the authors consider a gradient-free method to avoid using gradients due to unreliability stemming from gradient saturation~\citep{Adebayo2018,Kindermans2019}. The importance coefficient of a feature map is computed as a slope that measures the difference between the posterior class score and the score obtained when turning off that feature map. That difference is then normalized.

\subsection{Critical Analysis}
\label{subsec:critics}

Our review of several works on WSOL carried out in recent years showed the emergence of different strategies for WSOL tasks. We cite two main families. Top-down methods, which aim to back-trace the posterior probability class to find ROIs. These methods rely either on biology processes, classifier confidence, or gradients. Gradient-based methods, which are the more dominant. They are model-independent and can be used for any trained network, in addition to being easy to implement. This family of methods are also used for CNN visualization, and interpretability~\citep{Samek2019}.

While top-down methods have been successful, they have experienced slower progress than bottom-up methods, which seem to be the driving core of WSOL. Most successful WSOL methods derive from this family. They are easy to implement and follow the standard flow of information in feed-forward networks. Early works aimed to design different spatial pooling functions to obtain CAMs, but these methods quickly hit a fundamental snag  in CAMs, in the form of under-activation. This also suggests that relying only on spatial pooling to transfer global labels to the pixel-level is not enough. Subsequent works have focused on this issue mainly by attempting to recover the localization of a complete object. To that end, several strategies have been proposed:

\noindent \textbf{- Perturbation} of input images or embeddings, \ie, intermediate features. It is often used to mine discriminative regions. The most common perturbation mechanism is suppression, in which a part of the signal is deleted stochastically either uniformly or via selective attention.

\noindent \textbf{- Self-attention and self-learning.} Training CNNs to localize objects using only global labels is an ill-posed problem~\citep{wan2018min}. However, thanks to the convolutional filter properties, common patterns can emerge within intermediate spatial feature maps. Researchers exploit this property to collect self-attention maps, which often focus on objects. This self-attention has been successfully used to \emph{guide} intermediate convolution layers to further focus on emerging ROI and filter out background and noisy features. In addition, most confident regions in a self-attention map have been used as  self-supervisory signals at the pixel level.

\noindent \textbf{- Shallow features} have long been known to hold useful but noisy localization information in supervised learning, such as in segmentation~\citep{Ronneberger-unet-2015}. It was not until recently, however, that shallow features began to be exploited as well in WSOL tasks, allowing further boost the localization performance.

\noindent \textbf{- Pseudo-annotation} provides a substitute for full supervision. Using only global labels  for localization been somewhat very successful. They allow the identification of the most discriminative regions but are unable to recover full objects. Partial, noisy, and uncertain pseudo-supervision is currently deemed to be very useful to boost localization performance. It provides low-cost supervision, and yet, should be used with care since it could be very noisy, which could push the model in the wrong direction or trap it in local solutions that predict similar pseudo-annotations.

\noindent \textbf{- Decoupling classification and localization tasks.} Recent studies in WSOL have shown that these tasks are antagonistic: localization task converges during the very early epochs, and then later degrades, while the classification task converges toward the end of the training. Some works separate them by first training a classifier, then a localizer. The aim is to build a final framework that can yield the best performance over both tasks. Note that recent successful works on WSOL tasks combine shallow features with pseudo-annotations while separating classification and localization tasks.


\section{Experimental methodology}
\label{sec:methodology}

In this section, we present the experimental procedure used to evaluate the performance of deep WSOL models. The aim of our experiments was to assess and compare their ability to accurately classify histology images  and localize cancerous ROIs. 

In order to compare the localization performance of WSOL techniques on histology data, we selected different representative methods from both categories (bottom-up and top-down). From the \emph{bottom-up} category, we consider the following methods:
GAP~\citep{lin2013network}, 
MAX-Pool~\citep{oquab2015object}, 
LSE~\citep{sun2016pronet}, 
CAM~\citep{zhou2016learning}, 
HaS~\citep{SinghL17}, 
WILDCAT~\citep{durand2017wildcat}, 
ACoL~\citep{ZhangWF0H18}, 
SPG~\citep{ZhangWKYH18}, 
Deep MIL~\citep{ilse2018attention},
PRM~\citep{ZhouZYQJ18PRM}, 
ADL~\citep{ChoeS19}, 
CutMix~\citep{YunHCOYC19}, 
TS-CAM~\citep{gao2021tscam}, 
MAXMIN~\citep{belharbi2022minmaxuncer}, 
NEGEV~\citep{negevsbelharbi2022}; 
while the following methods are considered from the \emph{top-down} category: 
GradCAM~\citep{SelvarajuCDVPB17iccvgradcam}, 
GradCAM++~\citep{ChattopadhyaySH18wacvgradcampp},
Smooth-GradCAM++~\citep{omeiza2019corr}, 
XGradCAM~\citep{fu2020axiom}, 
LayerCAM~\citep{JiangZHCW21layercam}. 

Experiments are conducted on two public datasets of histology images, described in \autoref{subsec:datasets}. Most of the public datasets used are collected exclusively for classification or segmentation purposes~\citep{Daisuke201834}, these include the BreaKHis~\citep{spanhol-breakhis2016} and BACH~\citep{aresta2018bach} datasets. The only dataset we found that contained both image-level and pixel-level annotations was \glas (\autoref{subsubsec:glas2015}). Using a single dataset for evaluation could be insufficient to draw meaningful conclusions. Therefore, we created an additional dataset with the required annotations by using a protocol to sample image patches from WSIs of the CAMELYON16 dataset (\autoref{subsubsec:camelyon16}).

\subsection{Protocol}
\label{subsec:protocol}

In all our experiments, we follow the same experimental protocol as found in~\citep{choe2020evaluating} which defines a clear setup to evaluate ROI localization obtained by a weakly supervised classifier. The protocol includes two main elements, namely, \emph{model selection}, and an \emph{evaluation metric} at the pixel level.

In a weakly supervised setup, model selection is critical. The learning scenario considered in our experiments entails two main tasks: Classification and localization, which are shown to be antagonistic tasks~\citep{belharbi2022fcam,choe2020evaluating}. While the localization task converges during the very early training epochs, the classification task converges at late epochs. Therefore, to yield a better localization model, an adequate model selection protocol is required. Following~\citep{choe2020evaluating}, and considering a full validation set labeled only at the global level, we randomly select a few samples to be labeled additionally at the pixel level. In particular, we select a few samples per class to yield a balanced set. These samples are used for model selection using localization measures. This selection is referred to as a \beloc selection. Model selection using a full validation set employing only classification measures with global labels is referred to as \becl selection. We provide results on both selection methods to assess their impact on performance. All results are reported using \beloc selection unless specified otherwise. In the next section, we present the evaluation metrics.

\subsection{Performance measures}
\label{subsec:metrics}
For each task, \ie, classification and localization, we consider their respective metric.

\subsubsection{Classification task}
We use a standard classification accuracy \cl,
  \begin{equation}
    \label{eq:eqx0}
    \cl = 100 \times \;\frac{\text{\#correctly classified samples}}{\text{\#samples}} \; (\%) \; ,
\end{equation}
where \emph{${\text{\#correctly classified samples}}$} is the total number of correctly classified samples, and \emph{${\text{\#samples}}$} is the total number of samples.

\subsubsection{Localization task}
The aim of WSOL is to produce a score map that is used to localize an object. In order to measure the quality of localization of ROIs, we consider the same protocol used in~\citep{choe2020evaluating}. Using the class activation map ${\bm{S}}$ of a target class, a binary map is obtained through thresholding. At pixel location ${(i, j)}$, this map is compared to the ground true mask ${\bm{T}}$. Following~\citep{choe2020evaluating}, we threshold the score map at $\tau$ to generate the binary mask $\{(i,j)\mid s_{ij}\geq \tau\}$. We consider the following localization metrics:

\textbf{\pxap}: We use the \pxap metric, presented in~\citep{choe2020evaluating}, which measures the pixel-wise precision-recall. At a specific threshold ${\tau}$, the pixel precision and recall are defined as,
{\small
\begin{align}
    \text{\pxprec}(\tau)=
    \frac{|\{s^{(n)}_{ij}\geq\tau\}\cap\{T^{(n)}_{ij}=1\}|}%
    {|\{s^{(n)}_{ij}\geq\tau\}|} \label{eq:pxprec} \;, \\
    \text{\pxrec}(\tau)=
    \frac{|\{s^{(n)}_{ij}\geq\tau\}\cap\{T^{(n)}_{ij}=1\}|}%
    {|\{T^{(n)}_{ij}=1\}|} \label{eq:pxrec} \;. 
\end{align}
}%
The \pxap metric marginalizes the threshold ${\tau}$ over a predefined set of thresholds\footnote{In all the experiments, we used ${\tau} \in [0, 1]$ with a step of ${0.001}$ as in~\citep{choe2020evaluating}.},  
\begin{equation}
\label{eq:pxap}
  \text{\pxap}:=\sum_l \text{\pxprec}(\tau_l) \times (\text{\pxrec}(\tau_l)-\text{\pxrec}(\tau_{l-1}))  \;,
\end{equation}
which is the area under the curve of the pixel precision-recall curve.

\textbf{\cmtx}: Since we are dealing with a medical application, it is important to assess true positives/negatives and false positives/negatives performance at the pixel level in order to have real insights into localization accuracy. Such information is not explicitly provided via the \pxap metric. Therefore, we consider measuring the confusion matrix by marginalizing the threshold ${\tau}$ similarly to what is done in the  \pxap metric. First, we compute each normalized component of the confusion matrix with respect to a fixed threshold as follows,
{\small
\begin{align}
    \text{\tp}(\tau)=
    \frac{|\{s^{(n)}_{ij}\geq\tau\}\cap\{T^{(n)}_{ij}=1\}|}%
    {|\{T^{(n)}_{ij}= 1\}|}  \label{eq:tp} \;, \\
    \text{\fn}(\tau)=
    \frac{|\{s^{(n)}_{ij} < \tau\}\cap\{T^{(n)}_{ij}=1\}|}%
    {|\{T^{(n)}_{ij}= 1\}|}  \label{eq:fn} \;, \\
    \text{\fp}(\tau)=
    \frac{|\{s^{(n)}_{ij} \geq \tau\}\cap\{T^{(n)}_{ij}=0\}|}%
    {|\{T^{(n)}_{ij}= 0\}|}  \label{eq:fp} \;, \\
    \text{\tn}(\tau)=
    \frac{|\{s^{(n)}_{ij} < \tau\}\cap\{T^{(n)}_{ij}=0\}|}%
    {|\{T^{(n)}_{ij}=0\}|} \label{eq:tn}\;, 
\end{align}
}%
where ${\tp, \fn, \fp, \tn}$ are the true positives, false negatives, false positives, and true negatives, respectively. Each component can be represented as a graph with the x-axis as the threshold ${\tau}$. Similarly to \pxap, we marginalize confusion matrix components over ${\tau}$ by measuring the area under each component, which is also the average since the step between thresholds is fixed. We report the percentage values of each component,
{\small
\begin{align}
    \text{\pxtp}:= 100 \times \sum_l \text{\tp}(\tau_l) \times (\tau_l - \tau_{l-1})  \label{eq:pxtp} \;, \\
    \text{\pxfn}:= 100 \times \sum_l \text{\fn}(\tau_l) \times (\tau_l - \tau_{l-1})  \label{eq:pxfn} \;, \\
    \text{\pxfp}:= 100 \times \sum_l \text{\fp}(\tau_l) \times (\tau_l - \tau_{l-1})  \label{eq:pxfp} \;, \\
    \text{\pxtn}:= 100 \times \sum_l \text{\tn}(\tau_l) \times (\tau_l - \tau_{l-1})  \label{eq:pxtn} \;.
\end{align}
}%

\subsection{Datasets}
\label{subsec:datasets}
In this section, we describe the two public datasets of histology images used in our experiments: \glas for colon cancer, and \camsixteen for breast cancer.

\begin{figure}[h!]
    \centering
    \includegraphics[width=0.4\linewidth]{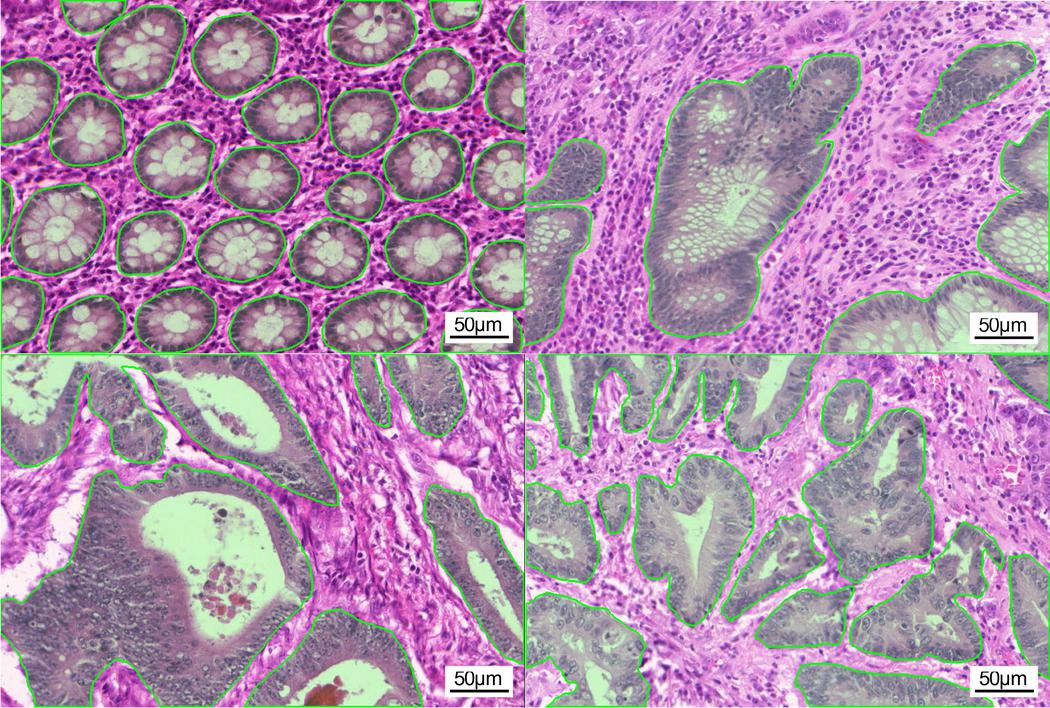}
    \caption{Example of images of different classes with their segmentations from \glas dataset (Credit:~\citep{sirinukunwattana2017gland}). \emph{Row 1}: Benign. \emph{Row 2}: Malignant.}
    \label{fig:fig4}
\end{figure}

\subsubsection{GlaS dataset (\glas)}
\label{subsubsec:glas2015}

This is a histology dataset for colon cancer diagnosis~\citep{sirinukunwattana2017gland}\footnote{The Gland Segmentation in Colon Histology Contest: \href{https://warwick.ac.uk/fac/sci/dcs/research/tia/glascontest}{https://warwick.ac.uk/fac/sci/dcs/research/tia/glascontest}}. It contains 165 images from 16 Hematoxylin and Eosin (H\&E) histology sections and their corresponding labels. For each image, both pixel-level and image-level annotations for cancer grading (\ie, benign or malign) are provided. The whole dataset is split into training (67 samples), validation (18 samples), and testing (80 samples) subsets. Among the validation set, 3 samples per class are selected to be fully supervised, \ie, 6 samples in total for \beloc selection.

\subsubsection{Camelyon16 dataset (\camsixteen)}
\label{subsubsec:camelyon16}

\begin{figure}[h!]
    \centering
    \includegraphics[width=0.4\linewidth]{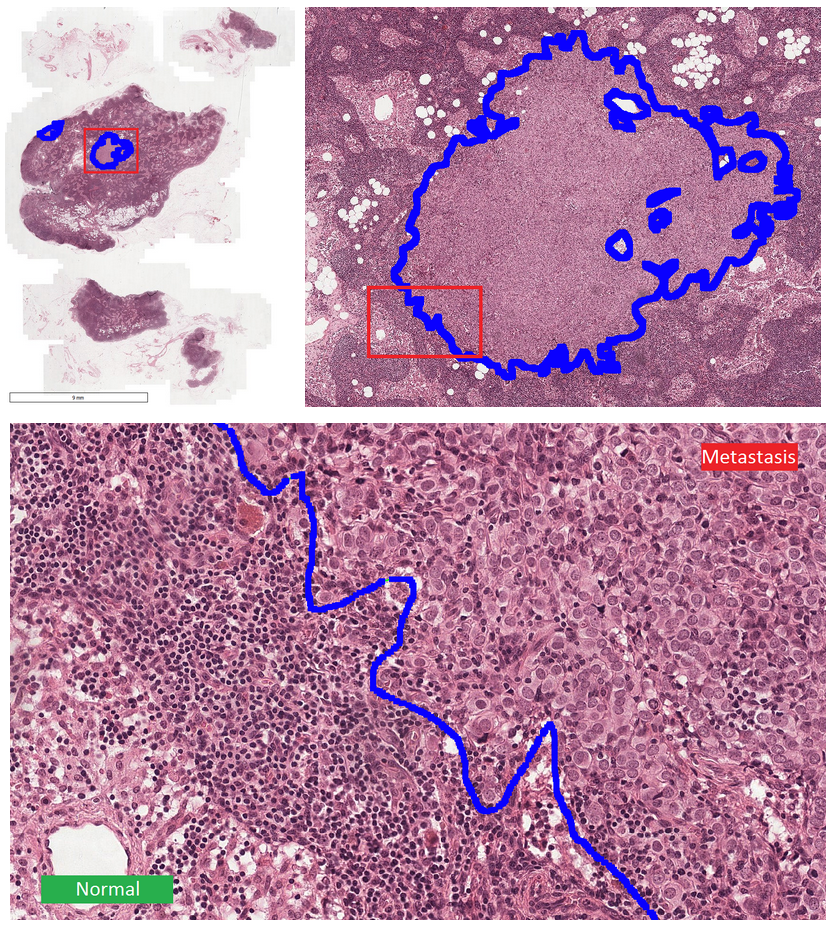}
    \caption{Example of metastatic regions in a WSI from CAMELYON16 dataset (Credit:~\citep{sirinukunwattana2017gland}). \emph{Top left}: WSI with tumor. \emph{Top right}: Zoom-in of one of the metastatic regions. \emph{Bottom}: Further zoom-in of the frontier between normal and metastatic regions.}
    \label{fig:fig5}
\end{figure}

\begin{figure}[h!]
  \center
\includegraphics[width=.4\linewidth]{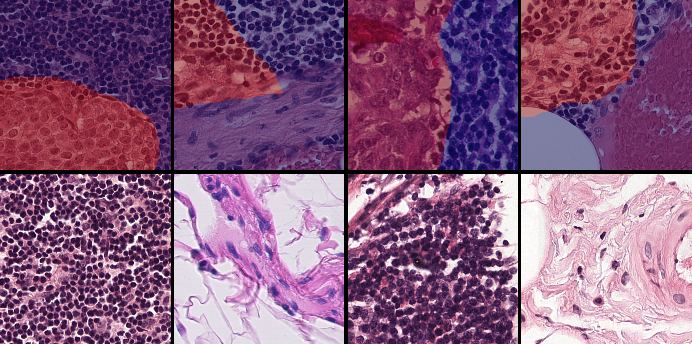}
\caption{Examples of test images from metastatic (top) and normal (bottom) classes of \camsixteen dataset of size ${512\times 512}$. Metastatic regions are indicated with a red mask.}
\label{fig:cam16-samples-patches512}
\end{figure}

This dataset\footnote{The Cancer Metastases in Lymph Nodes Challenge 2016 (CAMELLYON16): \href{https://camelyon16.grand-challenge.org/Home/}{https://camelyon16.grand-challenge.org/Home}} is composed of 399 WSI for detection of metastases in H\&E stained tissue sections of sentinel auxiliary lymph nodes (SNLs) of women with breast cancer \citep{camelyon2016paper}. The WSIs are annotated globally as  normal or metastases. The WSIs with metastases are further annotated at the pixel level to indicate regions of tumors. An example of a WSI is provided in \autoref{fig:fig5}. Among the 399 WSIs provided, 270 are used for training, and 129 for testing\footnote{Sample \texttt{test\_114} is discarded since the pixel level annotation was not provided. Therefore, the test set is composed of $128$ samples with $48$ samples with nodal metastases.}. The large size of the images makes their use in this survey inconvenient. Therefore, we designed a concise protocol to sample small sub-images for WSL with pixel-wise and image-level annotations. In summary, we sample sub-images of size ${512\times 512}$ to form  train, validation, and test sets, respectively (Fig.\ref{fig:cam16-samples-patches512}). A detailed sampling protocol is provided in \autoref{sec:sampling-camelyon16-tech-details}. 
This protocol generates a benchmark containing a total of 48,870 samples: 24,348 samples for training, 8,858 samples for validation, and 15,664 samples for testing. Each sub-set has balanced classes. For \beloc, we randomly select 5 samples per class from the validation set to be fully supervised, \ie, 10 samples in total.

{
\setlength{\tabcolsep}{3pt}
\renewcommand{\arraystretch}{1.1}
\begin{table}[ht!]
\centering
\resizebox{1.\textwidth}{!}{%
\centering
\small
\begin{tabular}{lc*{3}{c}gc*{3}{c}g}
& & \multicolumn{4}{c}{\glas} & & \multicolumn{4}{c}{\camsixteen}  \\
&  & VGG & Inception & ResNet & Mean &  & VGG & Inception & ResNet & Mean  \\
\cline{3-6}\cline{8-11} \\
Methods / Metric  &  & \multicolumn{9}{c}{\pxap (\beloc)}  \\
\cline{1-1}\cline{3-11} \\
\textbf{Bottom-up WSOL}  &  & \multicolumn{9}{c}{}  \\
\cline{1-1}\cline{3-6}\cline{8-11} \\
GAP~\citep{lin2013network} {\small \emph{(corr,2013)}}  &  & $58.5$  & $57.5$  & $56.2$  & 57.4  &  & $37.5$ & $24.6$ & $43.7$  & 35.2\\
MAX-Pool~\citep{oquab2015object} {\small \emph{(cvpr,2015)}}&  & $58.5$  & $57.1$  & $46.2$  & 53.9  &  & $42.1$& $40.9$ & $20.2$  & 34.4\\
LSE~\citep{sun2016pronet} {\small \emph{(cvpr,2016)}}  &  & $63.9$  & $62.8$  & $59.1$  & 61.9  &  & $63.1$ & $29.0$ & $42.1$  & 44.7\\
CAM~\citep{zhou2016learning} {\small \emph{(cvpr,2016)}}&  & $68.5$  & $50.5$  & $64.4$  & $ 61.1 $  &  & $25.4$ & $48.7$ & $27.5$  & 33.8\\
HaS~\citep{SinghL17} {\small \emph{(iccv,2017)}} &  & $65.5$  & $65.4$  & $63.5$  & 64.8  &  & $25.4$ & $47.1$ & $29.7$  & 34.0\\
WILDCAT~\citep{durand2017wildcat} {\small \emph{(cvpr,2017)}} &  & $56.1$  & $54.9$  & $60.1$ & 57.0  &  & $44.4$ & $31.4$ & $31.0$  & 35.6\\
ACoL~\citep{ZhangWF0H18} {\small \emph{(cvpr,2018)}} &  & $63.7$  & $58.2$  & $54.2$  & 58.7  &  & $31.3$ & $39.3$ & $31.3$  & 33.9\\
SPG~\citep{ZhangWKYH18} {\small \emph{(eccv,2018)}} &  & $63.6$  & $58.3$  & $51.4$  & 57.7  &  & $45.4$ & $24.5$ & $22.6$  & 30.8\\
Deep MIL~\citep{ilse2018attention} {\small \emph{(icml,2018)}}  &  & $66.6$  & $61.8$  & $64.7$  & 64.3  &  & $53.8$ & $51.1$ & $57.9$  & 54.2\\
PRM~\citep{ZhouZYQJ18PRM}  {\small \emph{(cvpr,2018)}}  &  & $59.8$  & $53.1$  & $62.3$  & $58.4$ &  & $46.0$ & $41.7$ & $23.2$  & $36.9$\\
ADL~\citep{ChoeS19} {\small \emph{(cvpr,2019)}} &  & $65.0$  & $60.6$  & $54.1$  & 59.9  &  & $19.0$ & $46.0$ & $46.0$  & 37.0\\
CutMix~\citep{YunHCOYC19} {\small \emph{(eccv,2019)}} &  & $59.9$  & $50.4$  & $56.7$  & 55.6  &  &  $56.4$ & $44.9$ & $20.7$  & 40.6\\
TS-CAM~\citep{gao2021tscam} {\small \emph{(corr,2021)}}  &  & t:$54.5$  & b:$57.8$  & s:$55.1$  & 52.8  &  & t:$46.3$ & b:$21.6$ & s:$42.2$  & 36.7\\
MAXMIN~\citep{belharbi2022minmaxuncer} {\small \emph{(tmi,2022)}}  &  & $75.0$  & $49.1$  & $81.2$  & 68.4  &  & $50.4$ & $\bm{80.8}$ & $\bm{77.7}$  & \textbf{69.6}\\
NEGEV~\citep{negevsbelharbi2022}  {\small \emph{(midl,2022)}}  &  & $\bm{81.3}$  & $\bm{70.1}$  & $\bm{82.0}$  & $\bm{77.8}$ &  & $\bm{70.3}$ & $53.8$ & $52.6$  & $58.9$\\
\cline{1-1}\cline{3-6}\cline{8-11} \\
\textbf{Top-down WSOL}  &  & \multicolumn{9}{c}{}  \\
\cline{1-1}\cline{3-6}\cline{8-11} \\
GradCAM~\citep{SelvarajuCDVPB17iccvgradcam} {\small \emph{(iccv,2017)}}  &  & $75.7$  & $56.9$  & $70.0$  & 67.5  &  & $40.2$ & $34.4$ & $29.1$  & 34.5\\
GradCAM++~\citep{ChattopadhyaySH18wacvgradcampp} {\small \emph{(wacv,2018)}} &  & $76.1$  & $65.7$  & $70.7$  & 70.8  &  & $41.3$ & $43.9$ & $25.8$  & 37.0\\
Smooth-GradCAM++~\citep{omeiza2019corr} {\small \emph{(corr,2019)}}  &  & $71.3$  & $67.6$  & $75.5$  & 71.4  &  & $35.1$ & $31.6$ & $25.1$  & 30.6\\
XGradCAM~\citep{fu2020axiom} {\small \emph{(bmvc,2020)}} &  & $73.7$  & $66.4$  & $62.6$  & 67.5  &  & $40.2$ & $33.0$ & $24.4$  & 32.5\\
LayerCAM~\citep{JiangZHCW21layercam} {\small \emph{(ieee,2021)}}  &  & $67.8$  & $66.1$  & $70.9$  & 68.2  &  & $34.1$ & $25.0$ & $29.1$  & 29.4\\
\cline{1-1}\cline{3-6}\cline{8-11} \\
\textbf{Fully supervised}  &  & \multicolumn{9}{c}{}  \\
\cline{1-1}\cline{3-6}\cline{8-11} \\
U-Net~\citep{Ronneberger-unet-2015}{\small \emph{(miccai,2015)}}  &  & $96.8$  & $95.4$  & $96.4$  & 96.2  &  & $83.0$ & $82.2$ & $83.6$  & 82.9\\
\cline{1-1}\cline{3-6}\cline{8-11} \\
\end{tabular}
}
\caption{\pxap performance over \glas and \camsixteen test sets. Model selection: \beloc.}
\label{tab:pxap-best-loc}
\vspace{-1em}
\end{table}
}

{
\setlength{\tabcolsep}{3pt}
\renewcommand{\arraystretch}{1.1}
\begin{table}[ht!]
\centering
\resizebox{1.\textwidth}{!}{%
\centering
\small
\begin{tabular}{lc*{4}{c}c*{4}{c}c*{4}{c}}
&  & \multicolumn{4}{c}{VGG} & & \multicolumn{4}{c}{Inception} & & \multicolumn{4}{c}{ResNet}\\
\cline{1-1}\cline{3-6}\cline{8-11}\cline{13-16} \\
\textbf{Bottom-up WSOL}  &  & \pxtp & \pxfn & \pxfp & \pxtn &  & \pxtp & \pxfn & \pxfp & \pxtn & & \pxtp & \pxfn & \pxfp & \pxtn   \\
\cline{1-1}\cline{3-6}\cline{8-11}\cline{13-16} \\
GAP~\citep{lin2013network} {\small \emph{(corr,2013)}}  &  & $34.5$  & $65.4$  & $20.5$  & 79.4  &  & $50.0$ & $49.9$ & $51.1$  & 48.8 &  & $30.6$ & $69.3$ & $21.2$  & 78.7\\
MAX-Pool~\citep{oquab2015object} {\small \emph{(cvpr,2015)}}&  & $38.7$  & $61.2$  & $31.5$  & 68.4  &  & $41.2$& $58.7$ & $36.3$  & 63.6 &  & $50.1$ & $49.8$ & $55.7$  & 44.2\\
LSE~\citep{sun2016pronet} {\small \emph{(cvpr,2016)}}  &  & $52.0$  & $47.9$  & $41.4$  & 58.5  &  & $\bm{70.9}$& $\bm{29.0}$ & $63.7$  & 36.2 &  & $51.0$ & $48.9$ & $44.9$  & 55.0\\
CAM~\citep{zhou2016learning} {\small \emph{(cvpr,2016)}}&  & $34.5$  & $65.4$  & $20.5$  & 79.4  &  & $50.0$& $49.9$ & $51.1$  & 48.8 &  & $30.6$ & $69.3$ & $21.2$  & 78.7\\
HaS~\citep{SinghL17} {\small \emph{(iccv,2017)}} &  & $42.3$  & $57.6$  & $31.4$  & 68.5  &  & $26.4$& $73.5$ & $16.1$  & 83.8 &  & $36.0$ & $63.9$ & $27.5$  & 72.4\\
WILDCAT~\citep{durand2017wildcat} {\small \emph{(cvpr,2017)}} &  & $37.6$  & $62.3$  & $33.2$  & 66.7  &  & $42.3$& $57.6$ & $35.6$  & 64.3 &  & $\bm{74.3}$ & $\bm{25.6}$ & $68.8$  & 31.1\\
ACoL~\citep{ZhangWF0H18} {\small \emph{(cvpr,2018)}} &  & $28.3$  & $71.6$  & $11.1$  & 88.8  &  & $6.6$& $93.3$ & $4.6$  & 95.3 &  & $16.1$ & $83.8$ & $6.4$  & 93.5\\
SPG~\citep{ZhangWKYH18} {\small \emph{(eccv,2018)}} &  & $62.2$  & $37.7$  & $50.5$  & 49.4  &  & $58.8$& $41.1$ & $51.9$  & 48.0 &  & $47.1$ & $52.8$ & $47.0$  & 52.9\\
Deep MIL~\citep{ilse2018attention} {\small \emph{(icml,2018)}}  &  & $14.7$  & $85.2$  & $\bm{7.3}$  & \textbf{92.6}  &  & $8.9$& $91.0$ & $4.3$  & 95.6 &  & $10.2$ & $89.7$ & $\bm{4.0}$  & \textbf{95.9}\\
PRM~\citep{ZhouZYQJ18PRM}  {\small \emph{(cvpr,2018)}}  &  & $41.8$  & $58.1$  & $34.7$  & 65.2  &  & $61.1$& $38.8$ & $59.7$  & 40.2 &  & $37.3$ & $62.6$ & $29.9$  & 70.0\\
ADL~\citep{ChoeS19} {\small \emph{(cvpr,2019)}} &  & $41.3$  & $58.6$  & $30.2$  & 69.7  &  & $47.3$& $52.6$ & $38.5$  & 61.4 &  & $34.4$ & $65.5$ & $31.4$  & 68.5\\
CutMix~\citep{YunHCOYC19} {\small \emph{(eccv,2019)}} &  & $41.3$  & $58.6$  & $33.6$  & 66.3  &  & $38.9$& $61.0$ & $39.4$  & 60.5 &  & $31.3$ & $68.6$ & $28.1$  & 71.8\\
TS-CAM~\citep{gao2021tscam} {\small \emph{(corr,2021)}}  &  & t:$23.1$  & t:$76.8$  & t:$20.4$  & t:79.5  &  & b:$25.2$ & b:$74.7$ & b:$20.4$  & b:79.5&  & s:$30.3$ & s:$69.6$ & s:$26.5$  & s:73.4\\
MAXMIN~\citep{belharbi2022minmaxuncer} {\small \emph{(tmi,2022)}}  &  & $\bm{57.4}$  & $\bm{42.5}$  & $41.8$  & 58.1  &  & $43.0$& $56.9$ & $44.3$  & 55.6 &  & $56.0$ & $43.9$ & $38.6$  & 61.3\\
NEGEV~\citep{negevsbelharbi2022}  {\small \emph{(midl,2022)}}  &  & $52.3$  & $47.6$  & $42.5$  & 57.4  &  & $54.7$& $45.2$ & $48.9$  & 51.0 &  & $52.2$ & $47.7$ & $45.6$  & 54.3\\
\cline{1-1}\cline{3-6}\cline{8-11} \cline{13-16}\\
\textbf{Top-down WSOL}  &  & \multicolumn{9}{c}{}  \\
\cline{1-1}\cline{3-6}\cline{8-11} \cline{13-16}\\
GradCAM~\citep{SelvarajuCDVPB17iccvgradcam} {\small \emph{(iccv,2017)}}  &  & $28.3$  & $71.6$  & $11.1$  & 88.8  &  & $6.6$& $93.3$ & $4.6$  & 95.3 &  & $16.1$ & $83.8$ & $6.4$  & 93.5\\
GradCAM++~\citep{ChattopadhyaySH18wacvgradcampp} {\small \emph{(wacv,2018)}} &  & $30.6$  & $69.3$  & $13.4$  & 86.5  &  & $12.5$& $87.4$ & $5.1$  & 94.8 &  & $19.0$ & $80.9$ & $8.6$  & 91.3\\
Smooth-GradCAM++~\citep{omeiza2019corr} {\small \emph{(corr,2019)}}  &  & $31.3$  & $68.6$  & $17.0$  & 82.9  &  & $15.6$& $84.3$ & $5.7$  & 94.2 &  & $24.8$ & $75.1$ & $10.0$  & 89.9\\
XGradCAM~\citep{fu2020axiom} {\small \emph{(bmvc,2020)}} &  & $31.5$  & $68.4$  & $14.9$  & 85.0  &  & $13.0$& $86.9$ & $4.1$  & 95.8 &  & $11.8$ & $88.1$ & $5.7$  & 94.2\\
LayerCAM~\citep{JiangZHCW21layercam} {\small \emph{(ieee,2021)}}  &  & $35.4$  & $64.5$  & $21.6$  & 78.3  &  & $11.0$& $88.9$ & $\bm{3.5}$  & \textbf{96.4} &  & $18.2$ & $81.7$ & $8.1$  & 91.8\\
\cline{1-1}\cline{3-6}\cline{8-11} \cline{13-16}\\
\textbf{Fully supervised}  &  & \multicolumn{9}{c}{}  \\
\cline{1-1}\cline{3-6}\cline{8-11}\cline{13-16} \\
U-Net~\citep{Ronneberger-unet-2015}{\small \emph{(miccai,2015)}}  &  & $89.8$  & $10.1$  & $11.6$  & 88.3  &  & $86.9$& $13.0$ & $14.6$  & 85.3 &  & $87.0$ & $12.9$ & $12.4$  & 87.5\\
\cline{1-1}\cline{3-6}\cline{8-11}\cline{13-16} \\
\end{tabular}
}
\caption{Confusion matrix performance over \glas test set. Model selection: \beloc.}
\label{tab:mtx-conf-best-loc-glas}
\vspace{-1em}
\end{table}
}

{
\setlength{\tabcolsep}{3pt}
\renewcommand{\arraystretch}{1.1}
\begin{table}[ht!]
\centering
\resizebox{1.\textwidth}{!}{%
\centering
\small
\begin{tabular}{lc*{4}{c}c*{4}{c}c*{4}{c}}
&  & \multicolumn{4}{c}{VGG} & & \multicolumn{4}{c}{Inception} & & \multicolumn{4}{c}{ResNet}\\
\cline{1-1}\cline{3-6}\cline{8-11}\cline{13-16} \\
\textbf{Bottom-up WSOL}  &  & \pxtp & \pxfn & \pxfp & \pxtn &  & \pxtp & \pxfn & \pxfp & \pxtn & & \pxtp & \pxfn & \pxfp & \pxtn   \\
\cline{1-1}\cline{3-6}\cline{8-11}\cline{13-16} \\
GAP~\citep{lin2013network} {\small \emph{(corr,2013)}}  &  & $54.0$  & $45.9$  & $52.6$  & 47.3  &  & $95.8$& $4.1$ & $38.0$  & 61.9 &  & $63.3$ & $36.6$ & $52.6$  & 47.3\\
MAX-Pool~\citep{oquab2015object} {\small \emph{(cvpr,2015)}}&  & $\bm{94.5}$  & $\bm{5.4}$  & $95.8$  & 4.1  &  & $70.8$& $29.1$ & $56.5$  & 43.4 &  & $75.7$ & $24.2$ & $85.0$  & 14.9\\
LSE~\citep{sun2016pronet} {\small \emph{(cvpr,2016)}}  &  & $80.9$  & $19.0$  & $53.5$  & 46.4  &  & $52.2$& $47.7$ & $48.8$  & 51.1 &  & $87.4$ & $12.5$ & $76.2$  & 23.7\\
CAM~\citep{zhou2016learning} {\small \emph{(cvpr,2016)}}&  & $54.0$  & $45.9$  & $52.6$  & 47.3  &  & $\bm{95.8}$& $\bm{4.1}$ & $38.0$  & 61.9 &  & $63.3$ & $36.6$ & $52.6$  & 47.3\\
HaS~\citep{SinghL17} {\small \emph{(iccv,2017)}} &  & $54.0$  & $45.9$  & $52.6$  & 47.3  &  & $90.5$& $9.4$ & $36.1$  & 63.8 &  & $53.8$ & $46.1$ & $48.6$  & 51.3\\
WILDCAT~\citep{durand2017wildcat} {\small \emph{(cvpr,2017)}} &  & $84.0$  & $15.9$  & $48.5$  & 51.4  &  & $40.1$& $59.8$ & $16.2$  & 83.7 &  & $37.4$ & $62.5$ & $21.4$  & 78.5\\
ACoL~\citep{ZhangWF0H18} {\small \emph{(cvpr,2018)}} &  & $13.4$  & $86.5$  & $4.6$  & 95.3  &  & $14.1$& $85.8$ & $7.2$  & 92.7 &  & $7.0$ & $92.9$ & $3.8$  & 96.1\\
SPG~\citep{ZhangWKYH18} {\small \emph{(eccv,2018)}} &  & $79.6$  & $20.3$  & $55.0$  & 44.9  &  & $47.7$& $52.2$ & $46.9$  & 53.0 &  & $47.1$ & $52.8$ & $48.1$  & 51.8\\
Deep MIL~\citep{ilse2018attention} {\small \emph{(icml,2018)}}  &  & $28.5$  & $71.4$  & $9.7$  & 90.2  &  & $54.4$& $45.5$ & $25.3$  & 74.6 &  & $25.2$ & $74.7$ & $7.6$  & 92.3\\
PRM~\citep{ZhouZYQJ18PRM}  {\small \emph{(cvpr,2018)}}  &  & $94.4$  & $5.5$  & $36.5$  & 63.4  &  & $63.5$& $36.4$ & $40.6$  & 59.3 &  & $00.0$ & $100.0$ & $00.0$  & 100.0\\
ADL~\citep{ChoeS19} {\small \emph{(cvpr,2019)}} &  & $51.8$  & $48.1$  & $56.5$  & 43.4  &  & $82.5$& $17.4$ & $41.7$  & 58.2 &  & $\bm{94.5}$ & $\bm{5.4}$ & $35.9$  & 64.0\\
CutMix~\citep{YunHCOYC19} {\small \emph{(eccv,2019)}} &  & $79.3$  & $20.6$  & $52.5$  & 47.4  &  & $73.7$& $26.2$ & $49.4$  & 50.5 &  & $1.5$ & $98.4$ & $22.0$  & 77.9\\
TS-CAM~\citep{gao2021tscam} {\small \emph{(corr,2021)}}  &  & t:$92.8$  & t:$7.1$  & t:$38.5$  & t:61.4  &  & b:$31.5$ & b:$68.4$ & b:$34.1$  & b:65.8 &  & s:$87.0$ & s:$12.9$ & s:$38.5$  & s:61.4\\
MAXMIN~\citep{belharbi2022minmaxuncer} {\small \emph{(tmi,2022)}}  &  & $47.1$  & $52.8$  & $47.1$  & 52.8  &  & $78.9$& $21.0$ & $29.6$  & 70.3 &  & $62.1$ & $37.8$ & $14.5$  & 85.4\\
NEGEV~\citep{negevsbelharbi2022}  {\small \emph{(midl,2022)}}  &  & $21.1$  & $78.8$  & $\bm{4.2}$  & \textbf{95.7}  &  & $22.1$& $77.8$ & $5.9$  & 94.0 &  & $9.1$ & $90.8$ & $3.9$  & 96.0\\
\cline{1-1}\cline{3-6}\cline{8-11} \cline{13-16}\\
\textbf{Top-down WSOL}  &  & \multicolumn{9}{c}{}  \\
\cline{1-1}\cline{3-6}\cline{8-11} \cline{13-16}\\
GradCAM~\citep{SelvarajuCDVPB17iccvgradcam} {\small \emph{(iccv,2017)}}  &  & $13.4$  & $86.5$  & $4.6$  & 95.3  &  & $14.1$& $85.8$ & $7.2$  & 92.7 &  & $7.0$ & $92.9$ & $3.8$  & 96.1\\
GradCAM++~\citep{ChattopadhyaySH18wacvgradcampp} {\small \emph{(wacv,2018)}} &  & $70.6$  & $29.3$  & $43.6$  & 56.3  &  & $16.1$& $83.8$ & $\bm{5.3}$  & \textbf{94.6} &  & $4.6$ & $95.3$ & $\bm{2.9}$  & \textbf{97.0}\\
Smooth-GradCAM++~\citep{omeiza2019corr} {\small \emph{(corr,2019)}}  &  & $33.7$  & $66.2$  & $22.5$  & 77.4  &  & $14.7$& $85.2$ & $8.6$  & 91.3 &  & $34.0$ & $65.9$ & $31.8$  & 68.1\\
XGradCAM~\citep{fu2020axiom} {\small \emph{(bmvc,2020)}} &  & $13.4$  & $86.5$  & $4.6$  & 95.3  &  & $26.4$& $73.5$ & $16.0$  & 83.9 &  & $15.1$ & $84.8$ & $15.4$  & 84.5\\
LayerCAM~\citep{JiangZHCW21layercam} {\small \emph{(ieee,2021)}}  &  & $13.2$  & $86.7$  & $6.2$  & 93.7  &  & $25.7$& $74.2$ & $23.8$  & 76.1 &  & $5.9$ & $94.0$ & $3.7$  & 96.2\\
\cline{1-1}\cline{3-6}\cline{8-11} \cline{13-16}\\
\textbf{Fully supervised}  &  & \multicolumn{9}{c}{}  \\
\cline{1-1}\cline{3-6}\cline{8-11}\cline{13-16} \\
U-Net~\citep{Ronneberger-unet-2015}{\small \emph{(miccai,2015)}}  &  & $58.9$  & $41.0$  & $7.1$  & 92.8  &  & $54.6$& $45.3$ & $5.4$  & 94.5 &  & $58.7$ & $41.2$ & $8.1$  & 91.8\\
\cline{1-1}\cline{3-6}\cline{8-11}\cline{13-16} \\
\end{tabular}
}
\caption{Confusion matrix performance over \camsixteen test set. Model selection: \beloc.}
\label{tab:mtx-conf-best-loc-camelyon16}
\vspace{-1em}
\end{table}
}

\subsubsection{Implementation details}
\label{subsubsec:impl-details}
The training of all methods is performed using SGD with 32 batch size~\citep{choe2020evaluating}, 1000 epochs for \glas, and 20 epochs for \camsixteen. We use a weight decay of ${10^{-4}}$. Images are resized to 256x256, and patches of size 224x224 are randomly sampled for training. Since almost all methods were evaluated on natural images, we cannot use the reported best hyper-parameters in their original papers. For each method, we perform a search for the best hyper-parameter over the validation set, including the learning rate. For the methods~\citep{negevsbelharbi2022,belharbi2022minmaxuncer}, we set part of the hyper-parameters as described in their papers, since they were evaluated on the same histology datasets.
For each method, the number of hyper-parameters to tune ranges from one to six. We use three different common backbones~\citep{choe2020evaluating}: VGG16 \citep{SimonyanZ14a}, InceptionV3 \citep{SzegedyVISW16}, and ResNet50 \citep{heZRS16}. For the TS-CAM method~\citep{gao2021tscam}, we use DeiT-based architectures~\citep{touvron21a}: DeiT-Ti (t), DeiT-S (s), DeiT-B (b). 
We use U-Net~\citep{Ronneberger-unet-2015} with full pixel annotation to yield an upper-bound segmentation performance. The weights of all architectures (backbones) are initialized using pre-trained models over Image-Net~\citep{KrizhevskyNIPS2012}. Then, ann the weights are trained on the histology data. The U-Net decoder is initialized randomly.  In \autoref{sec:hyper-params-search}, we provide full details on the hyper-parameters search.

\begin{figure*}[h!]
  \center
\includegraphics[width=.8\linewidth]{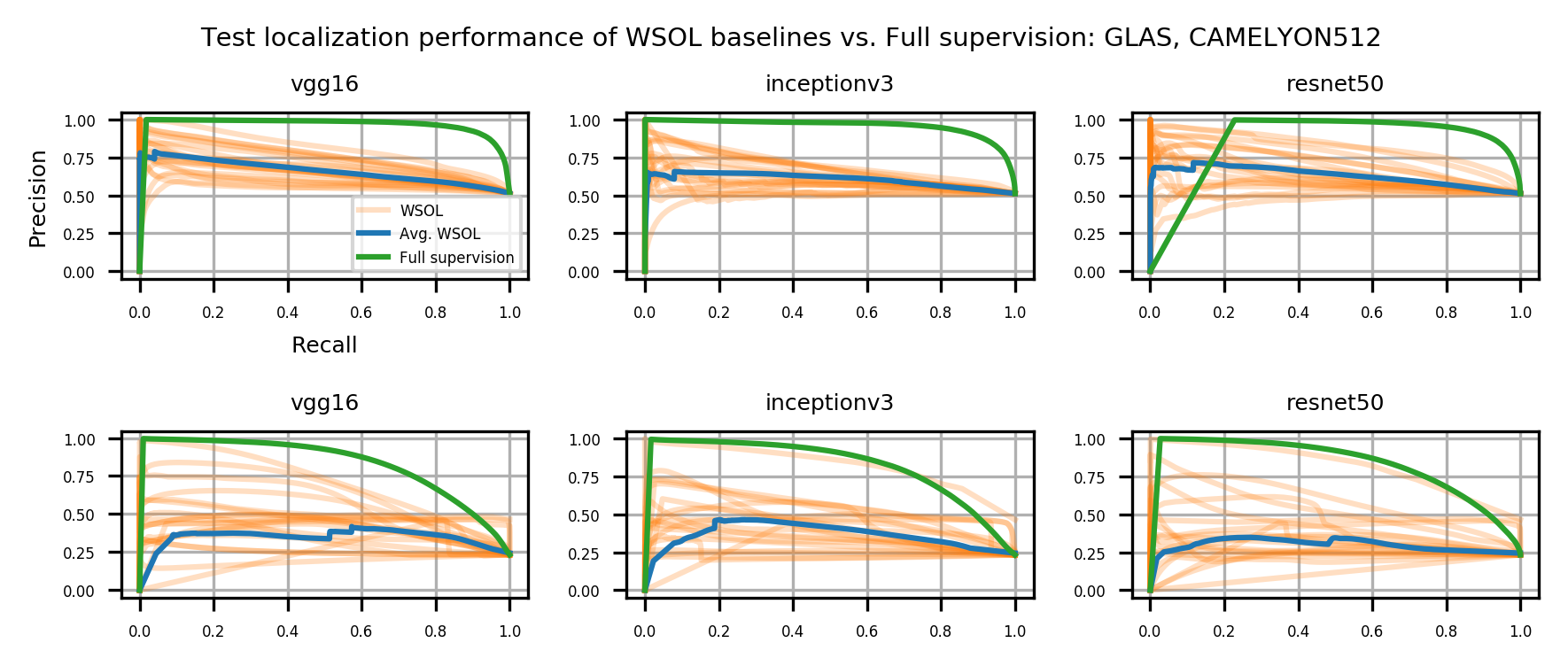}
\caption{Localization sensitivity to thresholding: WSOL methods (orange), average WSOL methods (blue), fully supervised method (green). Top: \glas. Bottom: \camsixteen. Best visualized in color.}
\label{fig:wsol-vs-fsup}
\end{figure*}

\section{Results and discussion}
\label{sec:results}

\subsection{Comparison of selected methods}
\label{subsec:main-results}

\autoref{tab:pxap-best-loc} shows the  localization performance (\pxap) of all methods over both the \glas and \camsixteen datasets. Overall, we observed a discrepancy in performance between different backbones. Across all WSOL methods, we obtained an average \pxap localization performance of ${66.86\%}$ for VGG, followed by ${63.46\%}$ for ResNet50, and finally,	${59.60\%}$ for Inception over the \glas test set (\autoref{tab:pxap-best-loc}). Over \camsixteen, VGG still ranks first with ${42.17\%}$, followed by Inception with ${40.61\%}$, and then ResNet50 with ${34.72\%}$. This performance difference comes from the basic architectural design difference between these common backbones.
In addition, the results of the WSOL methods show that the \camsixteen dataset, with an average localization performance of ${39.01\%}$, is more challenging than the \glas dataset, which has an average localization performance of ${62.75\%}$. This reflects the inherited difficulty in the \camsixteen dataset. While both datasets are challenging, the task in the \glas dataset boils down to localizing glands which often have a relatively distinct but variable shape or texture (\autoref{fig:fig4}). This makes spotting them relatively easy even for a non-expert. However, ROIs in \camsixteen have no obvious/common shape or texture (\autoref{fig:cam16-samples-patches512}). They can seem completely random from a local perspective. Spotting these ROIs can be even extremely challenging for non-experts. This explains the difference in the localization performance of WSOL methods over both datasets.
Note that methods that were designed for histology images such as MAXMIN~\citep{belharbi2022minmaxuncer} and NEGEV~\citep{negevsbelharbi2022} yield the best localization performance compared to generic methods that were designed and evaluated on natural images.  Top-down methods, such as GradCAM++~\citep{ChattopadhyaySH18wacvgradcampp} and LayerCAM~\citep{JiangZHCW21layercam}, have been shown to be more efficient on \glas, with an average of ${69.08\%}$, than bottom-up methods, with an average of ${60.65\%}$. However, bottom-up methods perform better on \camsixteen, with an average of ${41.09\%}$, as  compared to ${32.80\%}$ for top-down methods. This also can be explained by the aforementioned difference between both datasets. Bottom-up methods rely on convolution-responses that allow spotting common patterns,  which can be better detected in \glas on \camsixteen. However, top-down methods often rely on gradients that can spot arbitrary shapes giving these methods  more advantage over \camsixteen. The deep MIL method~\citep{ilse2018attention} yielded interesting results on both datasets.

Overall, the results also show a large performance gap between learning with weak supervision (global label in this case) and the fully-supervised method. This highlights the difficulty with histology images as  compared to natural images.

We also look into the confusion matrix to better assess the pixel-wise predictions on both the \glas (\autoref{tab:mtx-conf-best-loc-glas}) and \camsixteen (\autoref{tab:mtx-conf-best-loc-camelyon16}) datasets. The first observation is the high \emph{false negative} rate, with a large part of the ROI  considered as background. This metric goes up to ${\sim93\%}$  over \glas and ${100\%}$ over \camsixteen. This indicates that WSOL methods tend to under-activate by highlighting only a small part of the object and missing the rest. Under-activation is a common behavior in the WSOL method over natural images~\citep{choe2020evaluating}, which increases false negatives. We observe a new trend, namely, high \emph{false positive}, which is less common in WSOL~\citep{choe2020evaluating}. This is caused by the over-activation of the entire image, including the ROI and background. The visual similarity between foreground/background regions is the source of this issue, as the model is unable to discriminate between both regions. On average, false positives are much more frequent in \camsixteen than in \glas. However, in both datasets, the false negative rate is much higher than the false positive rate.

These results suggest that when dealing with histology images, the WSL method can exhibit two behaviors, either under-activate or over-activate, leading to high false negatives or positives. Both drawbacks should be considered when designing WSOL methods for this type of data. We provide visual results of both behaviors in \autoref{sec:visual-results}. In the histology literature, two different ways are considered to alleviate these issues. In~\citep{belharbi2022minmaxuncer}, the authors consider explicitly adding a background prior to prevent over-activation, while simultaneously preventing under-activation by promoting large sizes for both the background and the foreground. The authors in~\citep{negevsbelharbi2022} have considered using pixel-level guidance from a pre-trained classifier. Empirically, this allowed consistent patterns to emerge while avoiding under-/over-activation. However, the main drawback of this method is its strong dependence on the quality of pixel-wise evidence collected from the pre-trained classifier.

\subsection{Localization sensitivity to thresholds}
\label{subsec:loc-sensitivity-to-threshld}

In WSOL, thresholding is required~\citep{choe2020evaluating} to obtain a mask of a specific ROI. All the localization performances reported in this work are marginalized over a set of thresholds to allow a fair comparison~\citep{choe2020evaluating}. It has been shown that threshold values are critical for localization performance~\citep{choe2020evaluating}. In practice, for a test sample, one needs to threshold the CAM to yield a discrete localization of the ROI\footnote{Activations of CAMs can also be exploited visually by the user to determine ROI and manually inspect them.}. Ideally, the value of the threshold should not have a major impact on the ROI localization. However, that is not the case for WSOL. Evaluations on natural images~\citep{belharbi2022fcam,choe2020evaluating} have shown that localization performance from CAMs obtained in weakly-supervised setups is strongly tied to thresholds. We perform a similar analysis to the variation of localization performance with respect to the threshold in \autoref{fig:wsol-vs-fsup}. We observe a steep decline in performance when increasing the threshold, similar to the results obtained in~\citep{belharbi2022fcam,choe2020evaluating}. This once again highlights the dependency of localization on the threshold, and also indicates that optimal thresholds are concentrated near zero, similarly to the observation of~\citep{belharbi2022fcam,choe2020evaluating}. These results also suggest that CAM's activation distribution has a single mode located near zero as demonstrated  in~\citep{belharbi2022fcam,choe2020evaluating}. This makes finding an optimal threshold difficult, and therefore, makes CAMs  sensitive to thresholding, which reflects the uncertainty in CAMs. Ideally, the activation distribution is expected to be bimodal: background mode near zero, and foreground mode near one. Consequently, separating the foreground from the background becomes easy and less sensitive to the threshold, such as in a fully supervised method. The vulnerability of CAMs to thresholding is still an open issue in WSOL~\citep{belharbi2022fcam,choe2020evaluating}, and this should be considered in future designs of WSOL methods for general purposes, including histology data applications.

Results in \autoref{fig:wsol-vs-fsup} also show that there is still a large performance gap between WSOL and fully supervised methods, with the gap being much larger in the \camsixteen dataset.
\begin{figure}[hpt!]
  \center
  \subfloat[\textbf{Localization}: Impact of model selection (\beloc: orange. vs. \becl: blue) over test \textbf{localization} (\pxap) performance. Each point indicates the epoch (x-axis) at which the best model is selected and its corresponding localization performance (y-axis). Large circles indicate the average over all WSOL methods. Top: \glas. Bottom: \camsixteen.\label{fig:early-stopping-loc}]{
  \includegraphics[width=.9\linewidth]{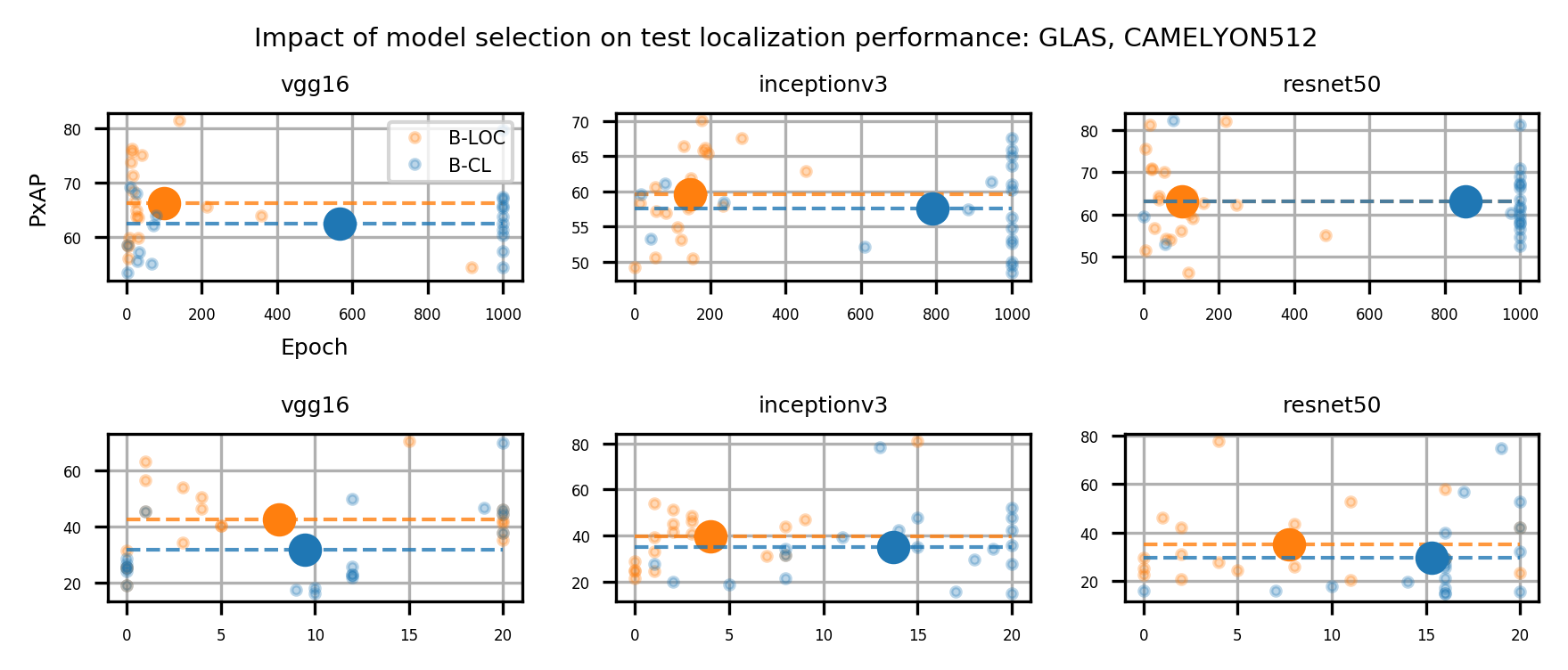}
  }\hfill
  \subfloat[\textbf{Classification}:  Impact of model selection (\beloc: orange. vs. \becl: blue) over test \textbf{classification} (\cl) performance. Each point indicates the epoch (x-axis) at which the best model is selected and its corresponding classification performance (y-axis). Large circles indicate the average over all WSOL methods. Top: \glas. Bottom: \camsixteen.\label{fig:early-stopping-cl}]{
  \includegraphics[width=.9\linewidth]{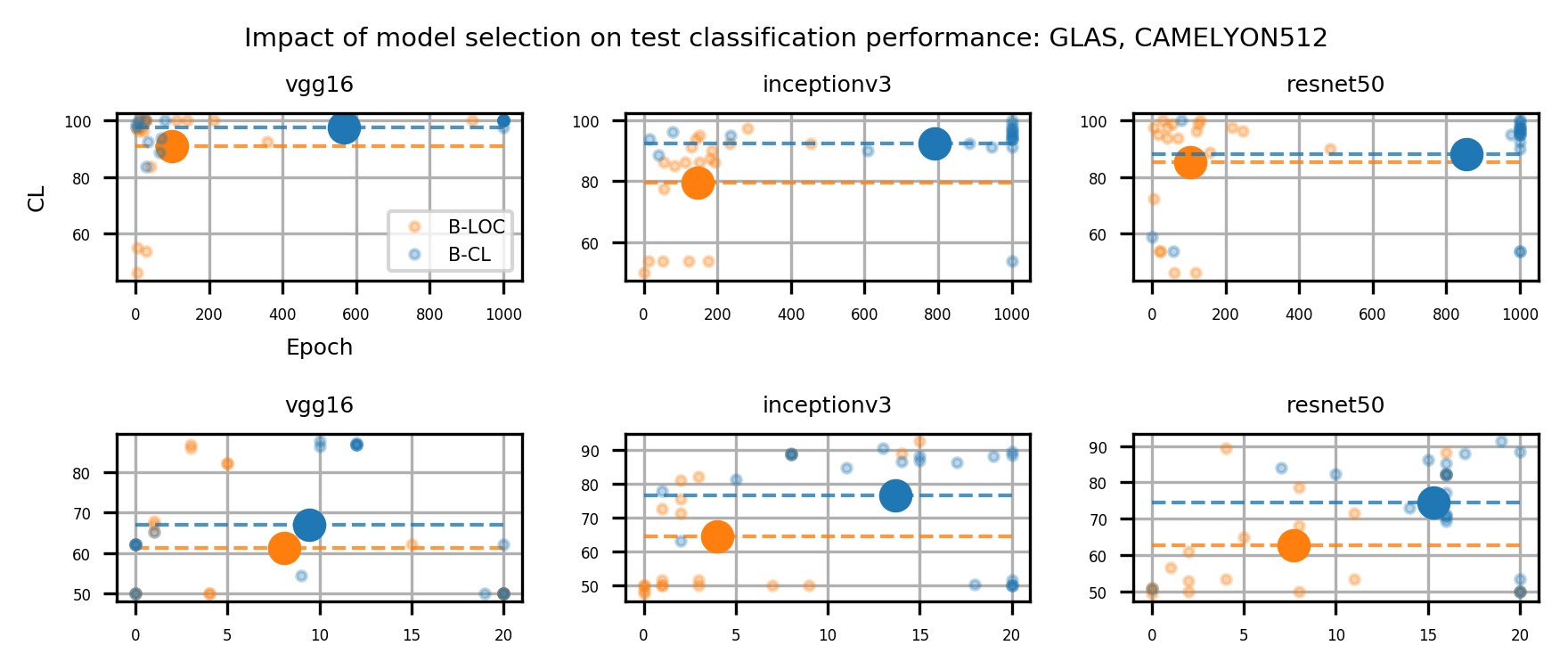}
  }\hfill
\caption{Impact of model selection (\beloc: orange. vs. \becl: blue) on test \textbf{localization} (\pxap) and \textbf{classification} (\cl) performance.}
\label{fig:early-stopping}
\end{figure}

\subsection{Importance of model selection}
\label{subsec:early-stopping-valid}

Typically, training a model for localization task with image class as weak supervision is done without having access to any localization information. This means that only classification information can be used to perform model selection using early stopping via a validation set, for instance.
However, it has been shown that classification and localization tasks are antagonistic~\citep{belharbi2022fcam,choe2020evaluating}. This implies that model selection via localization (\beloc) leads to good localization, and more likely than not, to poor classification.  On the other hand, selection using classification (\becl) yields good classification, and more likely than not, to poor localization, as observed empirically in~\citep{belharbi2022fcam,choe2020evaluating}.

The authors in~\citep{choe2020evaluating} suggested a standard protocol for WSOL, in which a few samples in the validation set are fully labeled, \ie, they bear localization annotations. Such subset is used for model selection using localization (\beloc) performance. While unrealistic\footnote{Since in a real weakly supervised application, we do not have any localization information}, this protocol allows a fair comparison between methods by removing user bias from the selection. 

In \autoref{fig:early-stopping-loc}, we show the impact of model selection on localization performance. We observe two main trends. First, localization performance converges during the very early epochs, while classification converges toward the end. This indicates that localization performance reaches its peak early on, and then degrades with long training epochs. The opposite can be said about classification. This result is consistent with the findings in~\citep{belharbi2022fcam,choe2020evaluating} over natural images.
A second important observation is that, on average, using localization information for model selection yields slightly better localization performance as compared to when using classification measures. Note that the total average gap varies on both datasets with ${\sim2\%}$ for \glas, and ${\sim8\%}$ for \camsixteen. This suggests that difficult datasets may benefit more from localization selection. Details of the differences in localization performance are reported in \autoref{tab:pxap-best-cl-vs-loc}.

In parallel, we inspected the impact of model selection on classification performance \autoref{fig:early-stopping-cl}. In most cases, we found that selecting a model based on its classification performance over a validation set consistently yields a largely better  model in terms of classification. In contrast, model selection using localization performance yields poor classification performance. From \ref{tab:classification-best-loc-cl}, \autoref{fig:early-stopping-cl}, in ${70.39\%}$ of the cases, the classification selection outperforms the localization selection, with a performance gap of: [min: ${0.10\%}$, max: ${54.3\%}$, average: ${15.08\%}$]. In addition, the reverse is true for ${11.18\%}$ of the cases with a performance gap of: [min: ${0.29\%}$, max: ${11.3\%}$ avg: ${4.45\%}$]. In ${18.42\%}$ of the cases, where both strategies yield the same classification performance. Therefore, based on these statistics, a better classifier is more likely to be selected using classification accuracy.

All these results mimic what is observed in~\citep{choe2020evaluating}, once again confirming that classification and localization under a weakly-supervised setup are more likely to be antagonistic. This is another challenge to consider when designing WSL methods.

The issue of model selection in WSOL was highlighted in~\citep{choe2020evaluating}, including its impact on localization and classification performance on natural images. A similar behavior is observed here on the histology dataset, with no clear solution. Most research works on WSOL aim primarily to improve the state-of-the-art localization perfromance. Works in~\citep{negevsbelharbi2022, belharbi2022fcam, ZhangCW20rethink} propose to separate localization from the classification task. First, they train a classifier to get the best classification performance. Then, using information from the classifier, a localizer is trained. This strategy allows to obtain a final framework that yields the best classification and localization performance.

\subsection{Computational complexity}
\label{subsec:inf-runtime}
Inference time is an important consideration for systems deployed in real-world applications. \autoref{tab:runetime} shows that WSOL methods require run times that are suitable for daily clinical routines. However, these statistics show that bottom-up methods are relatively faster than top-down methods. Note that bottom-up methods simply require a forward pass to yield the classification and localization. However, top-down methods require an additional backward pass to perform localization which, adds to the   processing time. Among top-down methods, confidence-based aggregation methods have proven to be very slow  since they need a series of forward passes to compute multiple scores. In addition, we observe that methods behave differently depending on the backbone architecture, with ResNet50 appearing to yield a slower performance than other backbones. \autoref{tab:n-parameters} presents more details about the memory resources required for backbones, including the number of parameters.

{
\setlength{\tabcolsep}{3pt}
\renewcommand{\arraystretch}{1.1}
\begin{table*}[ht!]
\centering
\resizebox{.8\linewidth}{!}{%
\centering
\small
\begin{tabular}{lc*{1}{c}c*{1}{c}c*{1}{c}}
 & & \multicolumn{5}{c}{\textbf{CNN Backbones}} \\
\cline{3-7} \\
\textbf{Methods}  & & \multicolumn{1}{c}{\textbf{VGG16}} & & \multicolumn{1}{c}{\textbf{Inception}}  & & \multicolumn{1}{c}{\textbf{ResNet50}} \\
\cline{1-1}\cline{3-3}\cline{5-5}\cline{7-7} \\
CAM~\citep{zhou2016learning} {\small \emph{(cvpr,2016)}} &  &     \multicolumn{1}{c}{0.2ms}  &&  \multicolumn{1}{c}{0.2ms}  &&  \multicolumn{1}{c}{0.3ms} \\
ACoL~\citep{ZhangWF0H18} {\small \emph{(cvpr,2018)}}  &  &                   \multicolumn{1}{c}{12.0ms}   && \multicolumn{1}{c}{19.2ms}  && \multicolumn{1}{c}{24.9ms} \\
SPG~\citep{ZhangWKYH18} {\small \emph{(eccv,2016)}} &  &                   \multicolumn{1}{c}{11.0ms}   && \multicolumn{1}{c}{18.0ms}  && \multicolumn{1}{c}{23.9ms} \\
ADL~\citep{ChoeS19} {\small \emph{(cvpr,2019)}} &  &    \multicolumn{1}{c}{6.4ms}  && \multicolumn{1}{c}{16.0ms}  && \multicolumn{1}{c}{14.4ms} \\
\shortstack{NEGEV}~\citep{negevsbelharbi2022} {\small \emph{(midl,2022)}}  &  &         \multicolumn{1}{c}{6.2ms}  && \multicolumn{1}{c}{25.5ms}  && \multicolumn{1}{c}{18.5ms} \\
\cline{1-1}\cline{3-3}\cline{5-5}\cline{7-7} \\
GradCAM~\citep{SelvarajuCDVPB17iccvgradcam} {\small \emph{(iccv,2017)}} &  &             \multicolumn{1}{c}{7.7ms}  && \multicolumn{1}{c}{21.1ms}  && \multicolumn{1}{c}{27.8ms} \\
GradCAM++~\citep{ChattopadhyaySH18wacvgradcampp} {\small \emph{(wacv,2018)}} &  &        \multicolumn{1}{c}{23.5ms}   && \multicolumn{1}{c}{23.7ms}  && \multicolumn{1}{c}{28.0ms}  \\
Smooth-GradCAM~\citep{omeiza2019corr} {\small \emph{(corr,2019)}} &  &                   \multicolumn{1}{c}{62.0ms}  && \multicolumn{1}{c}{150.7ms }  && \multicolumn{1}{c}{136.2ms} \\
XGradCAM~\citep{fu2020axiom} {\small \emph{(bmvc,2020)}} &  &                            \multicolumn{1}{c}{2.9ms} && \multicolumn{1}{c}{19.2ms}  && \multicolumn{1}{c}{14.2ms} \\
LayerCAM~\citep{JiangZHCW21layercam} {\small \emph{(ieee,2021)}} &  &                   \multicolumn{1}{c}{3.2ms}  && \multicolumn{1}{c}{18.2ms}   && \multicolumn{1}{c}{17.9ms}  \\
ScoreCAM~\citep{WangWDYZDMH20scorecam} {\small \emph{(cvpr,2020)}} &  &         \multicolumn{1}{c}{1.9s}   && \multicolumn{1}{c}{3.4s}   && \multicolumn{1}{c}{9.3s} \\
SSCAM~\citep{naidu2020sscam} {\small \emph{(corr,2020)}} && \multicolumn{1}{c}{105s}  && \multicolumn{1}{c}{136s}  &&\multicolumn{1}{c}{349s} \\
IS-CAM~\citep{naidu2020iscam} {\small \emph{(corr,2020)}} &  &  \multicolumn{1}{c}{30s}  && \multicolumn{1}{c}{39s}  && \multicolumn{1}{c}{99s} \\
\cline{1-1}\cline{3-3}\cline{5-5}\cline{7-7} \\
\end{tabular}
}
\caption{
Inference time required to produce CAMs using different WSOL methods with standard classifiers = encoder (VGG16, Inception, ResNet50) + global average pooling. The time needed to build a full-size CAM is estimated using an idle Tesla P100 GPU for one random RGB image of size ${224\times224}$ with ${200}$ classes. SSCAM~\citep{naidu2020sscam} (${N=35, \sigma=2}$), IS-CAM~\citep{naidu2020iscam} (${N=10}$), IS-CAM~\citep{naidu2020iscam} (${N=10}$) methods are evaluated with a batch size of 32 with their original hyper-parameters (${N, \text{ and } \sigma}$).
}
\label{tab:runetime}
\vspace{-1em}
\end{table*}
}

{
\setlength{\tabcolsep}{3pt}
\renewcommand{\arraystretch}{1.1}
\begin{table*}[ht!]
\centering
\resizebox{.5\linewidth}{!}{%
\centering
\small
\begin{tabular}{lc*{1}{c}c*{1}{c}c*{1}{c}}
 & & \multicolumn{5}{c}{\textbf{CNN Backbones}} \\
\cline{3-7} \\
\textbf{Measures}  & & \multicolumn{1}{c}{\textbf{VGG16}} & & \multicolumn{1}{c}{\textbf{Inception}}  & & \multicolumn{1}{c}{\textbf{ResNet50}} \\
 \cline{1-1}\cline{3-3}\cline{5-5}\cline{7-7} \\
\# parameters && $\approx$19.6M  && $\approx$25.6M  && $\approx$23.9M  \\
\# feature maps && 1024   &&  1024  && 2048 \\
\cline{1-1}\cline{3-7} \\
\end{tabular}
}
\caption{
The number of parameters per backbone, and the number of feature maps at the top layer. The size of feature maps at the top layer is ${28\times28}$.
}
\label{tab:n-parameters}
\vspace{-1em}
\end{table*}
}

{
\setlength{\tabcolsep}{3pt}
\renewcommand{\arraystretch}{1.1}
\begin{table}[ht!]
\centering
\resizebox{1.\textwidth}{!}{%
\centering
\small
\begin{tabular}{lc*{3}{c}gc*{3}{c}g}
& & \multicolumn{4}{c}{\glas} & & \multicolumn{4}{c}{\camsixteen}  \\
&  & VGG & Inception & ResNet & Mean &  & VGG & Inception & ResNet & Mean  \\
\cline{3-6}\cline{8-11} \\
Methods / Metric  &  & \multicolumn{9}{c}{\pxap: \beloc/\becl}  \\
\cline{1-1}\cline{3-11} \\
\textbf{Bottom-up WSOL}  &  & \multicolumn{9}{c}{}  \\
\cline{1-1}\cline{3-6}\cline{8-11} \\
GAP~\citep{lin2013network} {\small \emph{(corr,2013)}}  &  & \green{$58.5/53.5$}  & \red{$57.5/61.0$}  & \green{$60.3/56.2$}  & \green{57.4/56.9}  &  & $37.5/37.5$ & \green{$24.6/15.8$} & \green{$43.7/39.9$}  & \green{35.2/31.0}\\
MAX-Pool~\citep{oquab2015object} {\small \emph{(cvpr,2015)}}&  & $58.5/58.5$  & \green{$57.1/56.2$} & \red{$46.2/59.6$}  & \red{53.9/58.1}  &  & \green{$42.1/28.3$} & \green{$40.9/35.1$} & \green{$20.2/19.4$}  & \green{34.4/27.6}\\
LSE~\citep{sun2016pronet} {\small \emph{(cvpr,2016)}}  &  & \green{$63.9/62.2$}  & \green{$62.8/61.3$}  & \green{$59.1/58.1$}  & \green{61.9/60.5}  &  & \green{$63.1/25.7$} & \green{$29.0/27.9$} & \green{$42.1/32.0$}  & \green{44.7/28.5}\\
CAM~\citep{zhou2016learning} {\small \emph{(cvpr,2016)}}&  & \green{$68.5/60.3$} & \red{$50.5/53.3$}  & \green{$64.4/63.4$}  & \green{61.1/58.2}  &  & $25.4/25.4$ & \green{$48.7/39.4$} & \green{$27.5/14.8$}  & \green{33.8/26.5}\\
HaS~\citep{SinghL17} {\small \emph{(iccv,2017)}} &  & \green{$65.5/61.4$}  & \green{$65.4/63.6$}  & \green{$63.5/59.9$}  & \green{64.8/61.6}  &  & \green{$25.4/16.0$} & \red{$47.1/47.8$} & \green{$29.7/17.7$}  & \green{34.0/27.1}\\
WILDCAT~\citep{durand2017wildcat} {\small \emph{(cvpr,2017)}} &  & \red{$56.1/69.1$}  & \green{$54.9/48.4$}  & \green{$60.1/56.5$} & \red{57.0/58.0}  &  & $44.4/44.4$ & \red{$31.4/35.7$} & \green{$31.0/16.8$}  & \green{35.6/32.3}\\
ACoL~\citep{ZhangWF0H18} {\small \emph{(cvpr,2018)}} &  & \green{$63.7/57.2$}  & \green{$58.2/54.8$}  & \green{$54.2/53.0$}  & \green{58.7/55.0}  &  & \green{$31.3/18.1$} & \red{$39.3/42.2$} & \red{$31.3/32.0$}  & \green{33.9/30.7}\\
SPG~\citep{ZhangWKYH18} {\small \emph{(eccv,2018)}} &  & \green{$63.6/55.7$}  & \red{$58.3/59.5$}  & \red{$51.4/61.3$}  & \red{57.7/58.8}  &  & $45.4/45.4$ & \green{$24.5/14.9$} & \green{$22.6/15.5$}  & \green{30.8/25.2}\\
Deep MIL~\citep{ilse2018attention} {\small \emph{(icml,2018)}}  &  & \green{$66.6/63.7$}  & \green{$61.8/57.4$}  & \green{$64.7/57.9$}  & \green{64.3/59.6}  &  & \green{$53.8/49.8$} & \green{$51.1/47.9$} & \green{$57.9/56.9$}  & \green{54.2/51.5}\\
PRM~\citep{ZhouZYQJ18PRM}  {\small \emph{(cvpr,2018)}}  &  & \green{$59.8/57.4$}  & \green{$53.1/52.1$}  & \green{$62.3/58.8$}  & \green{58.4/56.1} &  & $46.0/46.0$ & \green{$41.7/21.6$} & \green{$23.2/16.0$}  & \green{36.9/27.8}\\
ADL~\citep{ChoeS19} {\small \emph{(cvpr,2019)}} &  & \green{$65.0/62.5$}  & \green{$60.6/49.5$}  & \red{$54.1/61.8$}  & \green{59.9/57.9}  &  & $19.0/19.0$ & \green{$46.0/29.8$} & \green{$46.0/16.0$}  & \green{37.0/21.6}\\
CutMix~\citep{YunHCOYC19} {\small \emph{(eccv,2019)}} &  & \green{$59.9/55.2$}  & \green{$50.4/49.9$}  & \green{$56.7/52.5$}  & \green{55.6/52.5}  &  &  \green{$56.4/25.4$} & \green{$44.9/27.6$} & \green{$20.7/14.7$}  & \green{40.6/22.5}\\
TS-CAM~\citep{gao2021tscam} {\small \emph{(corr,2021)}}  &  & t:$54.5/54.5$  & \red{b:$57.8/58.4$}  & \green{s:$55.1/54.7$}  & \red{52.8/55.8}  &  & \green{t:$46.3/17.5$} & \green{b:$21.6/34.1$} & s:$42.2/42.2$  & \green{36.7/31.2}\\
MAXMIN~\citep{belharbi2022minmaxuncer} {\small \emph{(tmi,2022)}}  &  & \green{$75.0/63.8$}  & \red{$49.1/61.0$}  & \red{$81.2/82.3$}  & \red{68.4/69.0}  &  & \green{$50.4/46.6$} & \green{$80.8/78.1$} & \green{$77.7/74.9$}  & \green{69.6/66.5}\\
\cline{1-1}\cline{3-6}\cline{8-11} \\
\textbf{Top-down WSOL}  &  & \multicolumn{9}{c}{}  \\
\cline{1-1}\cline{3-6}\cline{8-11} \\
GradCAM~\citep{SelvarajuCDVPB17iccvgradcam} {\small \emph{(iccv,2017)}}  &  & \green{$75.7/65.8$}  & \green{$56.9/52.6$}  & \green{$70.0/67.2$}  & \green{67.5/61.8}  &  & \green{$40.2/22.7$} & $34.4/34.2$ & $29.1/29.1$  & \green{34.5/28.6}\\
GradCAM++~\citep{ChattopadhyaySH18wacvgradcampp} {\small \emph{(wacv,2018)}} &  & \green{$76.1/67.3$}  & \green{$65.7/53.1$}  & \green{$70.7/69.0$}  & \green{70.8/63.1}  &  & \green{$41.3/26.6$} & \green{$43.9/42.5$} & \red{$25.8/26.8$}  & \green{37.0/31.9}\\
Smooth-GradCAM++~\citep{omeiza2019corr} {\small \emph{(corr,2019)}}  &  & \green{$71.3/67.9$}  & \green{$67.6/67.5$}  & \green{$75.5/66.3$}  & \green{71.4/67.2}  &  & \green{$35.1/24.2$} & $31.6/31.6$ & \red{$25.1/25.3$}  & \green{30.6/27.0}\\
XGradCAM~\citep{fu2020axiom} {\small \emph{(bmvc,2020)}} &  & \green{$73.7/65.3$}  & \green{$66.4/60.2$}  & \red{$62.6/67.1$}  & \green{67.5/64.2}  &  & \green{$40.2/22.7$} & \green{$33.0/18.8$} & \green{$24.4/21.1$}  & \green{32.5/20.8}\\
LayerCAM~\citep{JiangZHCW21layercam} {\small \emph{(ieee,2021)}}  &  & \green{$67.8/67.0$}  & \green{$66.1/65.9$}  & $70.9/70.9$  & \green{68.2/67.9}  &  & \green{$34.1/21.8$} & \green{$25.0/20.1$} & $29.1/29.1$  & \green{29.4/23.6}\\
\cline{1-1}\cline{3-6}\cline{8-11} \\
\textbf{Fully supervised}  &  & \multicolumn{9}{c}{}  \\
\cline{1-1}\cline{3-6}\cline{8-11} \\
U-Net~\citep{Ronneberger-unet-2015}{\small \emph{(miccai,2015)}}  &  & $96.8$  & $95.4$  & $96.4$  & 96.2  &  & $83.0$ & $82.2$ & $83.6$  & 82.9\\
\cline{1-1}\cline{3-6}\cline{8-11} \\
\end{tabular}
}
\caption{Comparison of localization performance (\pxap) with respect to model selection method: \beloc/\becl over \glas and \camsixteen test sets.
Colors: \red{\cl (\beloc) < \cl (\becl)} means that localization performance \pxap obtained using \beloc is worse than that obtained using \becl. \green{\cl (\beloc) > \cl (\becl)} means that localization performance obtained using \beloc is better than that obtained using \becl. Better visualized with color. The NEGEV method~\citep{negevsbelharbi2022} is not considered in this table because the classifier is pre-trained and frozen. Only \beloc is used for model selection.}
\label{tab:pxap-best-cl-vs-loc}
\vspace{-1em}
\end{table}
}

{
\setlength{\tabcolsep}{3pt}
\renewcommand{\arraystretch}{1.1}
\begin{table}[ht!]
\centering
\resizebox{1.\linewidth}{!}{%
\centering
\small
\begin{tabular}{lc*{3}{c}gc*{3}{c}g}
& & \multicolumn{4}{c}{\glas} & & \multicolumn{4}{c}{\camsixteen}  \\
&  & VGG & Inception & ResNet & Mean &  & VGG & Inception & ResNet & Mean  \\
\cline{3-6}\cline{8-11} \\
Methods / Metric  &  & \multicolumn{9}{c}{\cl: \beloc/\becl}  \\
\cline{1-1}\cline{3-11} \\
\textbf{WSOL}  &  & \multicolumn{9}{c}{}  \\
\cline{1-1}\cline{3-6}\cline{8-11} \\
GAP~\citep{lin2013network} {\small \emph{(corr,2013)}}  &  & \red{$46.2/98.7$}  & \red{$93.7/96.2$}  & \red{$87.5/95.0$}  & \red{75.8/96.6}  &  & $50.0/50.0$ & \red{$50.0/86.3$} & \red{$68.1/69.2$}  & \red{56.0/68.5}\\
MAX-Pool~\citep{oquab2015object} {\small \emph{(cvpr,2015)}}&  & $97.5/97.5$  & \red{$86.2/93.7$}  & \red{$46.2/58.7$}  & \red{76.6/83.3}  &  & $50.0/50.0$& \red{$82.0/86.7$} & \red{$71.4/72.9$}  & \red{67.8/69.8}\\
LSE~\citep{sun2016pronet} {\small \emph{(cvpr,2016)}}  &  & \red{$92.5/93.7$}  & \green{$92.5/91.2$}  & \green{$100/96.2$}  & \green{95.0/93.7}  &  & \red{$67.8/86.7$} & \red{$50.0/77.8$} & \red{$61.0/86.2$}  & \red{59.6/83.5}\\
CAM~\citep{zhou2016learning} {\small \emph{(cvpr,2016)}}&  & $100/100$  & \red{$53.7/88.7$}  & \red{$97.5/98.7$}  & \red{83.7/95.8}  &  & $62.2/62.2$ & \red{$51.3/84.7$} & \red{$53.5/71.0$}  & \red{55.6/72.6}\\
HaS~\citep{SinghL17} {\small \emph{(iccv,2017)}} &  & \green{$100/97.5$}  & \red{$86.2/97.5$}  & \red{$93.7/100$}  & \red{93.3/98.3}  &  & \red{$62.2/87.5$} & $50.0/50.0$ & \red{$51.0/82.3$}  & \red{54.4/73.2}\\
WILDCAT~\citep{durand2017wildcat} {\small \emph{(cvpr,2017)}} &  & \red{$55.0/100$}  & \red{$86.2/95.0$}  & \red{$96.2/100$} & \red{79.1/98.3}  &  & $50.0/50.0$ & $50.0/50.0$ & \red{$50.0/77.0$}  & \red{50.0/59.0}\\
ACoL~\citep{ZhangWF0H18} {\small \emph{(cvpr,2018)}} &  & \green{$100/92.5$}  & \red{$95.0/96.2$}  & \red{$46.2/53.7$}  & \red{80.4/80.8}  &  & \red{$50.0/86.3$} & \red{$50.0/86.4$} & $50.0/50.0$  & \red{50.0/74.2}\\
SPG~\citep{ZhangWKYH18} {\small \emph{(eccv,2018)}} &  & \red{$53.7/83.7$}  & \red{$53.7/93.7$}  & \red{$72.5/97.5$}  & \red{59.9/91.6}  &  & $65.1/65.1$ & $50.0/50.0$ & \red{$49.4/88.5$}  & \red{54.8/67.8}\\
Deep MIL~\citep{ilse2018attention} {\small \emph{(icml,2018)}}  &  & \red{$96.2/100$}  & \red{$81.2/92.5$}  & \green{$98.7/95.0$}  & \red{92.0/95.8}  &  & \red{$86.6/87.0$} & \red{$71.3/88.0$} & \green{$88.1/87.8$}  & \red{82.0/87.6}\\
PRM~\citep{ZhouZYQJ18PRM}  {\small \emph{(cvpr,2018)}}  &  & \red{$96.2/100$}  & \red{$53.7/90.0$}  & \green{$96.2/92.5$}  & \red{82.0/94.1} &  & $50.0/50.0$ & \red{$75.5/88.6$} & $50.0/50.8$  & \red{58.5/63.1}\\
ADL~\citep{ChoeS19} {\small \emph{(cvpr,2019)}} &  & $100/100$  & \red{$77.5/93.7$}  & \red{$93.7/95.0$}  & \red{90.4/96.2}  &  & $50.0/50.0$ & \red{$50.0/50.1$} & \red{$56.6/84.1$}  & \red{52.2/61.4}\\
CutMix~\citep{YunHCOYC19} {\small \emph{(eccv,2019)}} &  & \green{$100/88.7$}  & \red{$86.2/95.0$}  & \green{$100/96.2$}  & \green{95.4/93.3}  &  &  \green{$66.8/62.2$} & \red{$80.8/88.2$} & \red{$53.0/70.3$}  & \red{66.8/73.5}\\
TS-CAM~\citep{gao2021tscam} {\small \emph{(corr,2021)}}  &  & t:$100/100$  & \red{b:$92.5/95.0$}  & s:$90.0/90.0$  & \red{94.1/95.0}  &  & \red{t:$50.0/54.4$} & \red{b:$48.3/88.1$} & s:$50.0/50.0$  & \red{49.4/64.1}\\
MAXMIN~\citep{belharbi2022minmaxuncer} {\small \emph{(tmi,2022)}}  &  & \red{$83.7/100$}  & \red{$50.0/96.2$}  & \red{$95.0/100$}  & \red{76.2/98.7}  &  & $50.0/50.0$ & \green{$92.4/90.3$} & \red{$89.2/91.3$}  & 77.2/77.2\\
\cline{1-1}\cline{3-6}\cline{8-11} \\
\textbf{Top-down WSOL}  &  & \multicolumn{9}{c}{}  \\
\cline{1-1}\cline{3-6}\cline{8-11} \\
GradCAM~\citep{SelvarajuCDVPB17iccvgradcam} {\small \emph{(iccv,2017)}}  &  & \red{$97.5/100$}  & \red{$85.0/93.7$}  & \green{$98.7/95.0$}  & \red{93.7/96.2}  &  & \red{$40.2/86.6$} & \red{$34.4/88.7$} & \red{$29.1/82.3$}  & \red{34.5/85.8}\\
GradCAM++~\citep{ChattopadhyaySH18wacvgradcampp} {\small \emph{(wacv,2018)}} &  & \red{$97.5/100$}  & \red{$87.5/100$}  & $53.7/53.7$  & \red{79.5/84.5}  &  & \red{$50.0/62.2$} & \red{$88.9/89.3$} & \red{$78.6/85.3$}  & \red{72.5/78.9}\\
Smooth-GradCAM++~\citep{omeiza2019corr} {\small \emph{(corr,2019)}}  &  & $100/100$  & \red{$97.5/98.7$}  & $97.5/97.5$  & \red{98.3/98.7}  &  & \red{$50.0/62.2$} & $88.5/88.5$ & \red{$51.0/82.3$}  & \red{63.1/77.6}\\
XGradCAM~\citep{fu2020axiom} {\small \emph{(bmvc,2020)}} &  & $100/100$  & $91.2/91.2$  & \red{$88.7/96.2$}  & \red{93.3/95.8}  &  & \red{$82.1/86.6$} & \green{$88.9/81.1$} & \green{$82.3/71.0$}  & \green{84.4/79.5}\\
LayerCAM~\citep{JiangZHCW21layercam} {\small \emph{(ieee,2022)}}  &  & $100/100$  & \red{$90.0/97.5$}  & $53.7/53.7$  & \red{81.2/83.7}  &  & \red{$85.8/86.6$} & \red{$47.4/62.9$} & $82.1/82.1$  & \red{71.7/77.2}\\
\cline{1-1}\cline{3-6}\cline{8-11} 
\end{tabular}
}
\caption{Comparison of classification accuracy (\cl) with different model selection method: \beloc/\becl on \glas and \camsixteen test sets. Colors: \red{\cl (\beloc) < \cl (\becl)} means that the classification accuracy obtained using \beloc is worse than when using \becl. \green{\cl (\beloc) > \cl (\becl)} means that  the classification accuracy obtained using \beloc is better than when using \becl. Better visualized with color. The NEGEV method~\citep{negevsbelharbi2022} is not considered in this table because the classifier (CAM~\citep{zhou2016learning}) is pre-trained and frozen. Only \beloc is used for model selection.} 
\label{tab:classification-best-loc-cl}
\vspace{-1em}
\end{table}
}


\section{Conclusion and future directions}
\label{sec:conclusion}

Training deep models for ROI localization in histology images requires costly dense annotation. In addition, such labeling is performed by medical experts. A weakly supervised object localization framework provides different techniques for low-cost training of deep models. Using only image-class annotation, WSOL methods can be trained to classify an image and yield a localization of ROIs via CAMs. Despite its success, the WSOL framework still faces a major challenge, namely, correctly transferring image-class labels to the pixel-level. Moreover, histology images present additional challenges over natural images, including their size, stain variation, and ambiguity of labels. Most importantly, ROIs in histology data are less salient, which makes spotting them much more difficult. This easily opens WSOL models up to false positives/negatives.

In this work, we have presented a review of several deep WSOL methods covering the  2013 to early 2022 period. We divided them into two main categories based on the information flow in the model: bottom-up, and top-down methods. The latter have seen limited progress, while bottom-up methods are the current driving force behind WSOL task. They have undergone several major changes which have greatly improved the task. Early works focused on designing different spatial pooling functions. However, these methods quickly peaked in term of performance, revealing a major limitation to CAMs, namely, under-activation. Subsequent works aimed to alleviate this issue and recover the complete object using different techniques including: perturbation, self-attention, shallow features, pseudo-annotation, and tasks decoupling. Recent state-of-the-art methods combine several of these techniques.

To assess the localization and classification performance of WSOL techniques over histology data, we selected representative methods in our taxonomy for experimentation. We evaluated them over two public histology datasets: one for colon cancer (\glas) and a second dataset for breast cancer (\camsixteen), using the standard protocol for a WSOL task~\citep{choe2020evaluating}. Overall, the results indicate poor localization performance, particularly for generic methods that were designed and evaluated over natural images. Methods designed considering histology data challenges yielded good results. In general, all methods suffer from high false positive/negative localization. Furthermore, our analysis showed the following issues:

\noindent\textbf{- Under-activation,} where CAMs activate only over a small discriminative region, which increases false negatives. This is a documented behavior over natural images.

\noindent \textbf{- Over-activation,} where CAMs activate over the entire image, and increases false positives. This is a new behavior of WSOL which is mostly caused by the similarity between foreground and background regions. This makes discrimination between both regions difficult. The following are common strategies to reduce both issues: 
\textbf{1)} Use of priors over the region size~\citep{belharbi2022minmaxuncer}. The size of both the foreground and background in an image is constrained to be as large as possible. However, the analytical solution to their constraint peaks when each region covers half of the image. Results showed that such a  trivial solution does not occur in practice due to the existence of other competing constraints. 
\textbf{2)} Use of pseudo-annotations~\citep{negevsbelharbi2022}. The authors used pixel-wise evidence for the foreground and background from a pre-trained classifier. This explicitly provides pixel-wise supervision to the model. Since the pre-trained classifier could easily yield wrong pseudo-labels, this makes the model vulnerable to learning from wrong labels and ties its performance to the performance of the pre-trained classifier. A better solution consists in building more reliable pseudo-labels. Synthesizing better substitution to full supervision via pseudo-annotation is a potential path to explore. Using noisy pseudo-labels has already shown interesting results over medical and natural data~\citep{songsurvey2020}.

\noindent \textbf{- Sensitivity to thresholding} which is another documented issue for CAMs over natural images, which was observed over histology data as well. A typical solution is to push CAMs' activation to be more confident. However, altering the distribution of CAMs could deteriorate the classification performance. A possible solution is to separate the classification scoring function from CAMs such as in the architecture proposed in~\citep{negevsbelharbi2022,belharbi2022fcam,tcamsbelharbi2023}.

\noindent \textbf{- Model selection.} Using only classification accuracy allowed to select models having high classification performance, but poor localization. On the other hand, using localization performance for model selection yielded the opposite. This is a common issue in the WSOL task. Recent works have suggested separating both tasks to obtain a framework that yields the best performance for both. The architecture proposed in~\citep{negevsbelharbi2022} represents a key solution to this issue. First, a classifier is trained, and then it is frozen. Next, the localizer is trained. When trained properly, the final model is expected to yield the best performance in both tasks.

Our final conclusion in this work is that the localization performance obtained with WSOL methods when  applied to histology data still lags behind performance with full supervision. The methods are still unable to accurately localize ROI, mainly due to their non-saliency. We have cited several key issues to be considered when designing future WSOL techniques for histology data in order to close the performance gap between weakly and fully supervised methods.


\acks{This research was supported in part by the Canadian Institutes of Health Research, the Natural Sciences and Engineering Research Council of Canada, and the Digital Research Alliance of Canada (alliancecan.ca).}

%
\ethics{The work follows appropriate ethical standards in conducting research and writing the manuscript, following all applicable laws and regulations regarding treatment of animals or human subjects.}

\coi{We declare we don't have any conflicts of interest.}

\bibliography{biblio}


\FloatBarrier

\appendix
\clearpage

\section{Hyper-parameter search}
\label{sec:hyper-params-search}
\autoref{tab:tabx-general-hyper-params} presents the general hyper-parameters used for all methods. \autoref{tab:tabx-per-method-hyper-params} holds hyper-parameters for specific methods.

\begin{table}[h!]
\renewcommand{\arraystretch}{1.3}
\caption{General hyper-parameters.}
\label{tab:tabx-general-hyper-params}
\centering
\resizebox{.7\linewidth}{!}{
\begin{tabular}{lccc}
    Hyper-parameter  &  Value  \\
    \toprule
    Fully sup. model f & U-Net\\
    \hline
    Backbones & VGG16, InceptionV3, ResNet50.\\
    \hline
    Optimizer &  SGD\\
    \hline
    Nesterov acceleration & True\\
    \hline
    Momentum & $0.9$ \\
    \hline
    Weight decay & $0.0001$\\
    \hline
    Learning rate & \makecell{${\in \{0.01, 0.001, 0.1\}}$} \\
    \hline
    \multirow{2}{*}{Learning rate decay} 
     & \glas: 0.1 each 250 epochs.\\
     & \camsixteen: 0.1 each 5 epochs. \\
     \hline
    Mini-batch size & \makecell{$32$} \\
    \hline
    Random flip & Horizontal/vertical random flip \\
    \hline
    \multirow{2}{*}{Random color jittering} 
    & $\mathrm{Brightness}$, $\mathrm{contrast}$, \\
    & and $\mathrm{saturation}$ at $0.5$ and $\mathrm{hue}$ at $0.05$ \\
    \hline
    \multirow{2}{*}{Image size} 
    & Resize image to $225\times 225$. \\
    & Then, crop random patches of $224\times 224$\\
    \hline
    Learning epochs & \makecell{\glas: $1000$, \camsixteen: $20$} \\
    \bottomrule
\end{tabular}
}
\end{table}

\begin{table}[h!]
\renewcommand{\arraystretch}{1.3}
\caption{Per-method hyper-parameters. Notation in this table follows the same notation in the original papers.}
\label{tab:tabx-per-method-hyper-params}
\centering
\resizebox{.7\linewidth}{!}{
\begin{tabular}{lccc}
    Hyper-parameter  &  Value  \\
    \toprule
    LSE~\citep{sun2016pronet} & ${q \in \{1, 2, 3, 4, 5, 6, 7, 8, 9, 10\}}$\\
    \hline
    \multirow{2}{*}{HaS~\citep{SinghL17}} & Grid size ${\in \{8, 16, 32, 44, 56\}}$, \\
    & Drop rate ${\in \{0.2, 0.3, 0.4, 0.5, 0.6\}}$\\
    \hline
    \multirow{4}{*}{WILDCAT~\citep{durand2017wildcat}} 
            & ${\alpha \in \{0.1, 0.6\}}$ \\
            & kmax ${\in \{0.1, 0.3, 0.5, 0.6, 0.7\}}$\\
            & kmin ${\in \{0.1, 0.2, 0.3\}}$\\
            & Modalities = 5\\
    \hline
    ACoL~\citep{ZhangWF0H18} & ${\delta \in \{0.5, 0.6, 0.7, 0.8, 0.9\}}$\\
    \hline
    \multirow{6}{*}{SPG~\citep{ZhangWKYH18}} 
            & ${\delta_{1h} \in \{0.5, 0.7\}}$ \\
            & ${\delta_{1l} \in \{0.01, 0.05, 0.1\}}$ \\
            & ${\delta_{2h} \in \{0.5, 0.6, 0.7\}}$ \\
            & ${\delta_{2l} \in \{0.01, 0.05, 0.1\}}$ \\
            & ${\delta_{3h} \in \{0.5, 0.6, 0.7\}}$ \\
            & ${\delta_{3l} \in \{0.01, 0.05, 0.1\}}$ \\
    \hline
    Deep MIL~\citep{ilse2018attention} & Mid-channels = 128. Gated attention: True/False.\\
    \hline
    \multirow{2}{*}{PRM~\citep{ZhouZYQJ18PRM}} 
            & ${r \in \{3, 5, 7, 9, 11, 13\}}$ \\
            & Kernel stride ${ \in \{1, 3, 5, 7, 9, 11, 13\}}$ \\
    \hline
    \multirow{2}{*}{ADL~\citep{ChoeS19}} 
            & Drop rate ${ \in \{0., 0.25, 0.35, 0.45, 0.50, 0.75\}}$ \\
            & ${\gamma \in \{0.75, 0.85, 0.90\}}$ \\
    \hline
    \multirow{2}{*}{CutMix~\citep{YunHCOYC19}} 
            & ${\alpha  = 1.0}$ \\
            & ${\lambda \in \{0.5, 0.6, 0.7, 0.8, 0.9, 1.\}}$ \\
    \hline
    MAXMIN~\citep{belharbi2022minmaxuncer} & Same as~\citep{belharbi2022minmaxuncer}\\
    \hline
    NEGEV~\citep{negevsbelharbi2022} & Same as~\citep{negevsbelharbi2022}\\
    \bottomrule
\end{tabular}
}
\end{table}

\section{CAMELYON16 protocol for WSL}
\label{sec:sampling-camelyon16-tech-details}

This appendix provides details on our protocol for creating a WSOL benchmark from the CAMELYON16 dataset \citep{camelyon2016paper}. Samples are patches from WSIs, and each patch has two levels of annotation:
\begin{itemize}[leftmargin=1em]
  \item Image-level label ${y}$: the class of the patch, where \\ ${{y} \in \{\texttt{normal}, \texttt{metastatic}\}}$. 
  \item Pixel-level label ${\bm{Y}=\{0, 1\}^{H^\text{in}\times W^\text{in}}}$: a binary mask where the value $1$ indicates a \texttt{metastatic} pixel, and $0$ a \texttt{normal} pixel. For \texttt{normal} patches, this mask will contain $0$ only.
\end{itemize}

First, we split the CAMELYON16 dataset into training, validation, and test sets at the \emph{WSI-level}. This prevents patches from the same WSI from ending up in different sets. All patches are sampled with the highest resolution from WSI --\ie, $\text{level}=0$ in WSI terminology--. Next, we present our methodology of sampling metastatic and normal patches.

\begin{description}[style=unboxed, leftmargin=0cm]
  \item[Sampling metastatic patches.] Metastatic patches are sampled only from metastatic WSIs around the cancerous regions. Sampled patches will have an image-level label, and a pixel-level label. The sampling follows these steps:
  \begin{enumerate}
    \item Consider a metastatic WSI.
    \item Sample a patch $\bm{x}$ with size $(H, W)$.
    \item Binarize the patch into a mask $\bm{x}^\text{b}$ using the OTSU method \citep{otsu1979}. Pixels with value $1$ indicate tissue.
    \item Let ${p^{\bm{x}^\text{b}}_t}$ be the tissue percentage within $\bm{x}^\text{b}$. If ${p^{\bm{x}^\text{b}}_t < p_t}$, discard the patch.
    \item Compute the metastatic binary mask $\bm{Y}$ of the patch $\bm{x}$ using the pixel-level annotation of the WSI (values of $1$ indicate a metastatic pixel).
    \item Compute the percentage $p^{\bm{x}}_m$ of metastatic pixels within  $\bm{Y}$.
    \item If $p^{\bm{x}}_m < p_0$, discard the patch. Else, keep the patch $\bm{x}$ and set ${y} = \texttt{metastatic}$ and ${\bm{Y}}$ is its pixel-level annotation.
  \end{enumerate}

  We note that we sample \emph{all} possible metastatic patches from CAMELYON16 using the above approach. Sampling using such an approach will lead to a large number of metastatic patches with a high percentage of cancerous pixels (patches sampled from the center of the cancerous regions). These patches will have their binary annotation mask $\bm{Y}$ full of 1s. Using these patches will shadow the performance measure of the localization of cancerous regions. To avoid this issue, we propose to perform a calibration of the sampled patches in order to remove most such patches. We define two categories of metastatic patches:
  \begin{enumerate}
    \item \mbox{\textbf{Category 1}}: Contains patches with ${p_0 \leq p^{\bm{x}}_m} \leq p_1$. Such patches are rare, and contain only a small region of cancerous pixels. They are often located at the edge of the cancerous regions within a WSI.
    \item \textbf{Category 2}: Contains patches with ${ p^{\bm{x}}_m} > p_1$. Such patches are extremely abundant, and contain a very large region of cancerous pixels (most often the entire patch is cancerous). Such patches are often located inside the cancerous regions within a WSI.
  \end{enumerate}
  
  Our calibration method consists in keeping all patches within \textbf{Category 1}, and discarding most of the patches in \textbf{Category 2}. To this end, we apply the following sampling approach:
  \begin{enumerate}
    \item Assume we have $n$ patches in \textbf{Category 1}. We will sample ${n \times p_n}$ patches from \textbf{Category 2}, where $p_n$ is a predefined percentage.
    \item Compute the histogram of the frequency of the percentage of cancerous pixels within all patches, assuming a histogram with $b$ bins.
    \item Among all the bins with $p^{\bm{x}}_m > p_1$, pick a bin uniformly.
    \item Pick a patch within that bin uniformly.
  \end{enumerate}
  
  This procedure is repeated until we sample ${n \cdot p_n}$ patches from \textbf{Category 2}. 

  In our experiments, patches are not overlapping. We use the following configuration: ${p_0 = 20\%}$, ${p_1 = 50\%}$, ${p_t = 10\%}$, ${p_n = 1 \%}$. The number of bins in the histogram is obtained by dividing the interval ${[0, 1]}$ with a delta of $0.05$. These hyper-parameters are not validated to optimize the performance of the models, but are set in a reasonable way to \emph{automatically} sample consistent and unbalanced patches without the need to manually check the samples. Patch size is set to $512\times 512$. \autoref{fig:figMetastaticPatches} illustrates an example of metastatic patches and their corresponding masks.

  \midsepremove
  \begin{figure}[htp!]
      \centering
      \includegraphics[width=.4\linewidth]{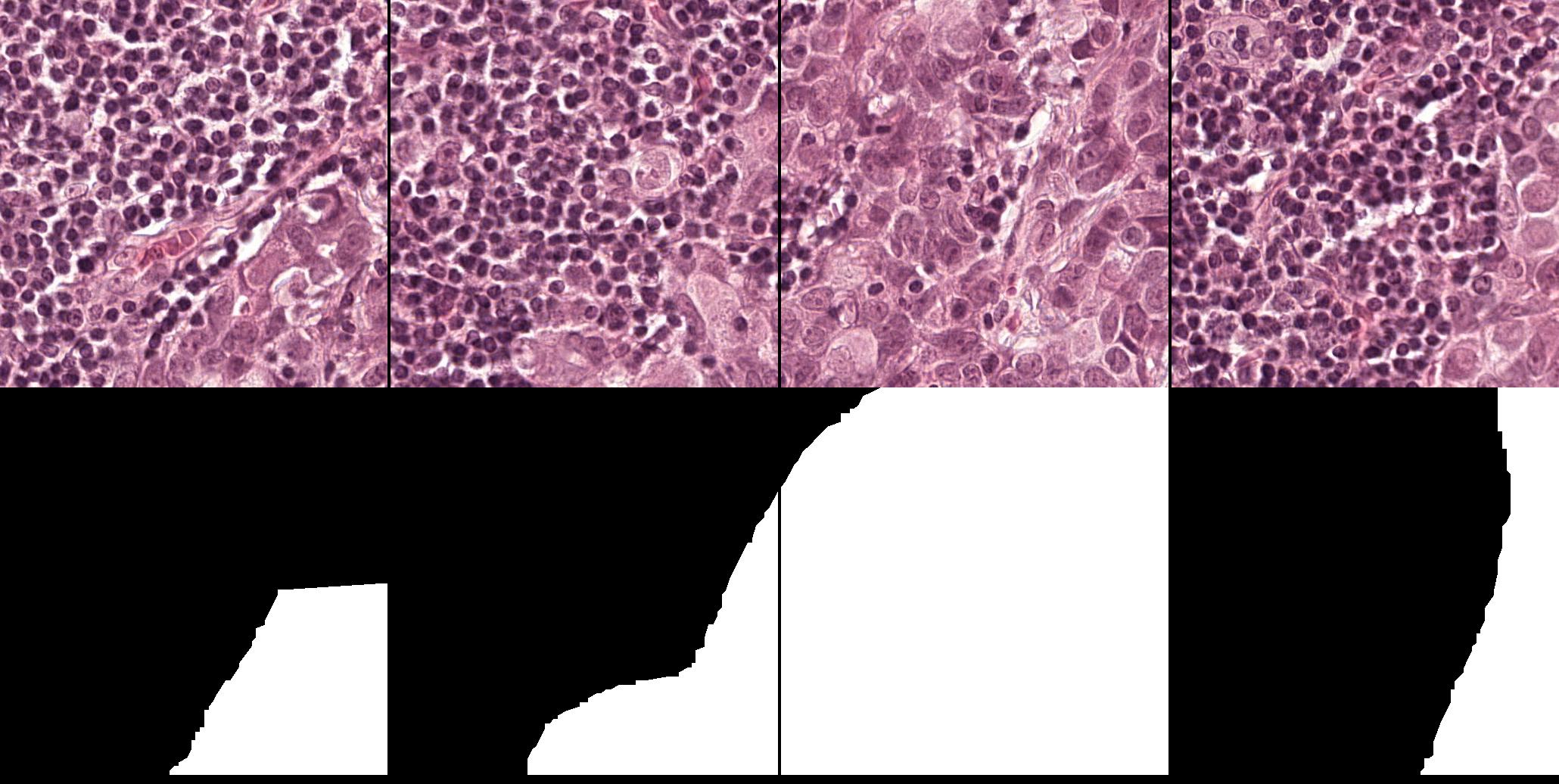}
      \caption{Example of metastatic patches with size $512\times 512$ sampled from CAMELYON16 dataset (WSI: \texttt{tumor\_001.tif}). \emph{Top row}: Patches. \emph{Bottom row}: Masks of metastatic regions (white color).}
      \label{fig:figMetastaticPatches}
  \end{figure}

  \midsepdefault
  
  \item[Sampling normal patches.] Normal patches are sampled only from normal WSI. A normal patch is sampled randomly and uniformly from the WSI (without repetition or overlapping). If the patch has enough tissue (${p^{\bm{x}^\text{b}}_t \geq p_t}$), the patch is accepted. Tissue mass measurement is performed at $\text{level}=6$ where it is easy for the OTSU binarization method to split the tissue from the background. We double-check the tissue mass at $\text{level}=0$.

  Let us consider a set (train, validation, or test) at patch level. We first pick the corresponding metastatic patches from the metastatic WSI, assuming $n_m$ is their total number. Assuming there are $h$ normal WSIs in this set, we sample the same number of normal patches as the total number of metastatic ones. In order to mix the patches from all the normal WSI, we sample ${\frac{n_m}{h}}$ normal patches per normal WSI. In our experiment, we use the same setup as in the metastatic patches sampling case: ${p_t = 10\%}$. \autoref{fig:figNormalPatches} illustrates an example of normal patches.

  \begin{figure}[htp!]
      \centering
      \includegraphics[width=.4\linewidth]{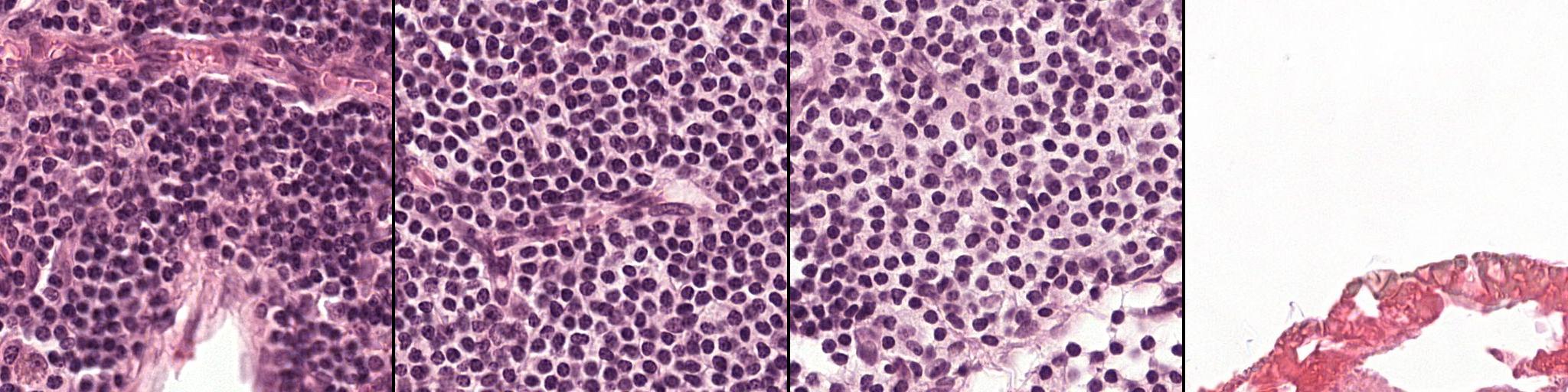}
      \caption{Example of normal patches with size $512\times 512$ sampled from CAMELYON16 dataset (WSI: \texttt{normal\_001.tif}).}
      \label{fig:figNormalPatches}
  \end{figure}
\end{description}

\section{Visual results}
\label{sec:visual-results}
\begin{figure*}[hbt!]
     \centering
         \centering
         \includegraphics[width=\linewidth]{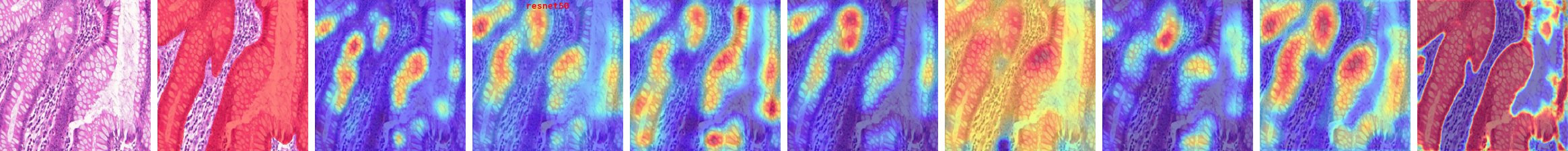}\\
         \includegraphics[width=\linewidth]{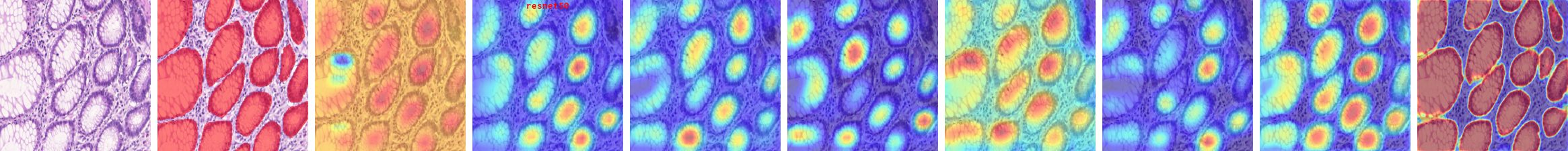}\\
         \includegraphics[width=\linewidth]{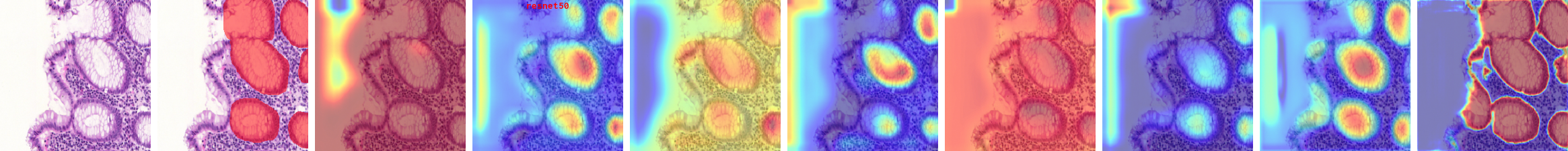}\\
         \includegraphics[width=\linewidth]{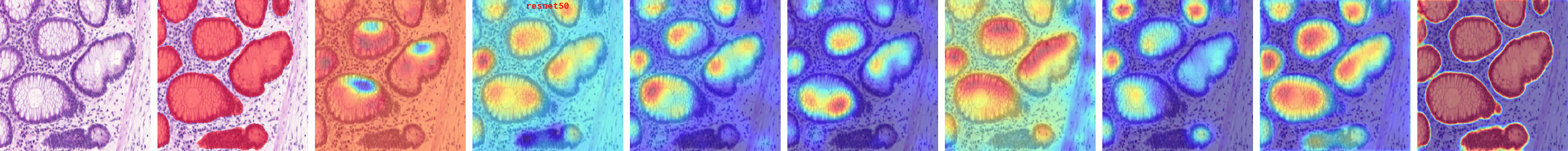}\\
         \includegraphics[width=\linewidth]{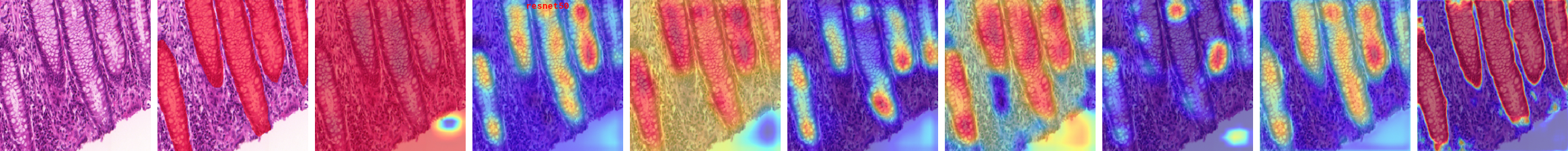}\\
         \includegraphics[width=\linewidth]{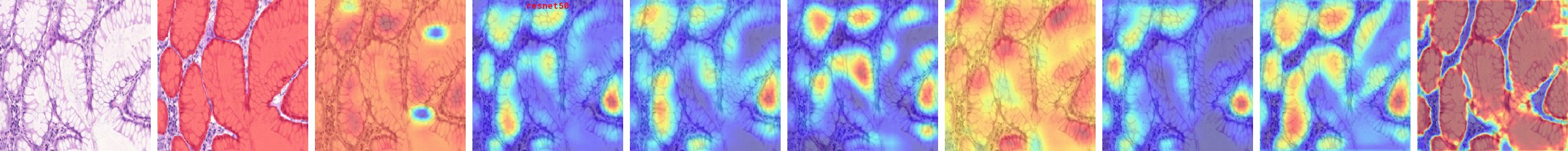}\\
         \includegraphics[width=\linewidth]{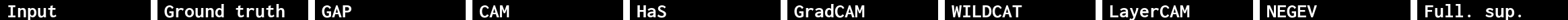}\\
        \caption{CAM predictions over \textbf{benign} test samples for \glas. Ground truth ROIs are indicated with a red mask highlighting glands. In all predictions, strong CAM's activations indicate glands. Backbone: ResNet50.}
        \label{fig:sup-glas-cams-benign}
\end{figure*}

\begin{figure*}[hbt!]
     \centering
         \centering
         \includegraphics[width=\linewidth]{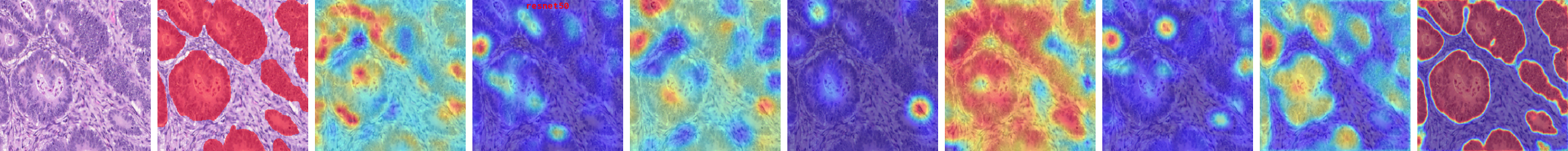}\\
         \includegraphics[width=\linewidth]{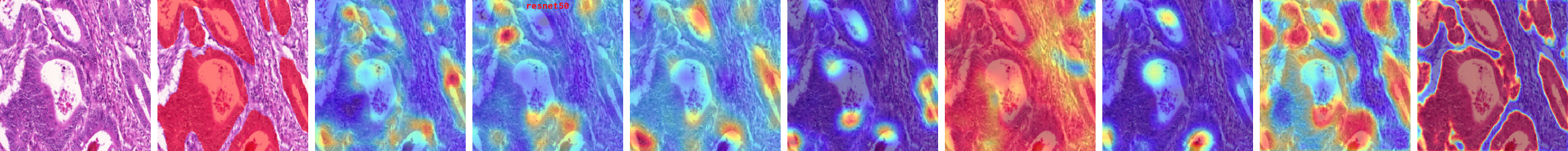}\\
         \includegraphics[width=\linewidth]{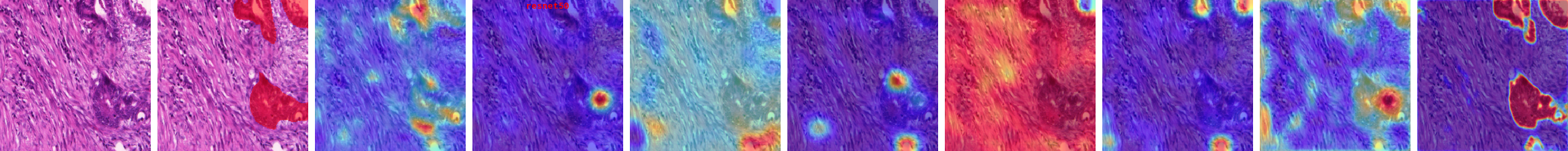}\\
         \includegraphics[width=\linewidth]{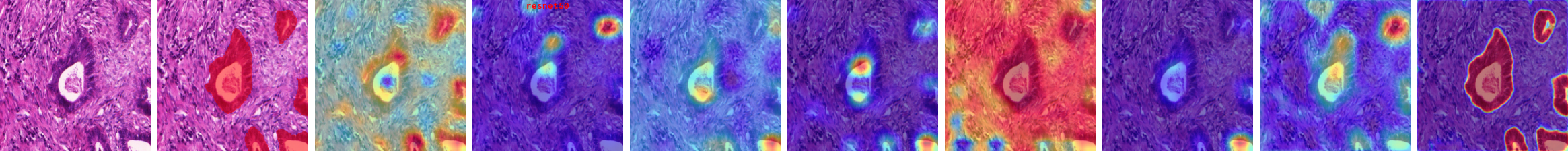}\\
         \includegraphics[width=\linewidth]{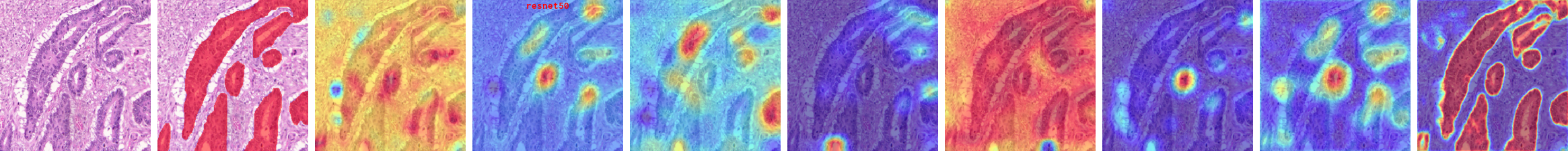}\\
         \includegraphics[width=\linewidth]{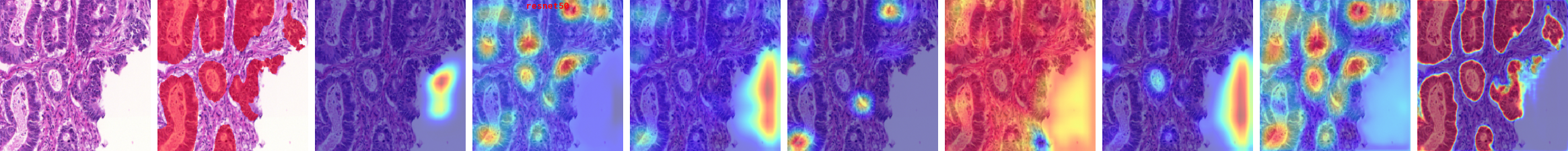}\\
         \includegraphics[width=\linewidth]{tag-GLAS}\\
        \caption{Predictions over \textbf{malignant} test samples for \glas. Ground truth ROIs are indicated with a red mask highlighting glands. In all predictions, strong CAM's activations indicate glands. Backbone: ResNet50.}
        \label{fig:sup-glas-cams-malignant}
\end{figure*}

\begin{figure*}[hbt!]
     \centering
         \centering
         \includegraphics[width=\linewidth]{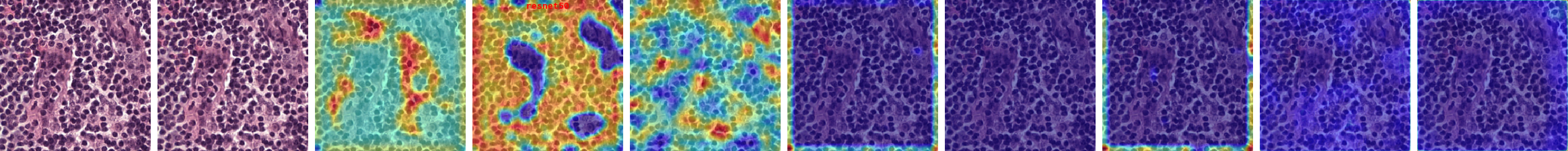}\\
         \includegraphics[width=\linewidth]{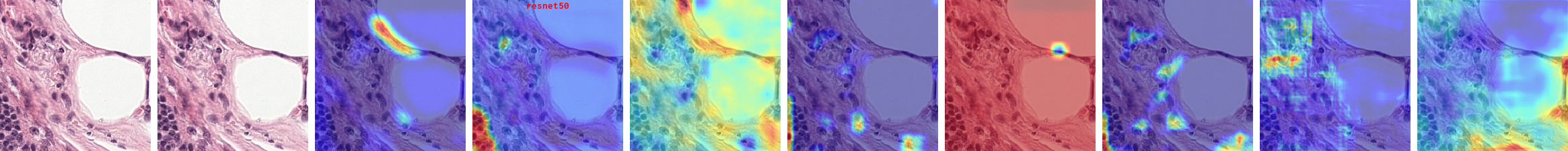}\\
         \includegraphics[width=\linewidth]{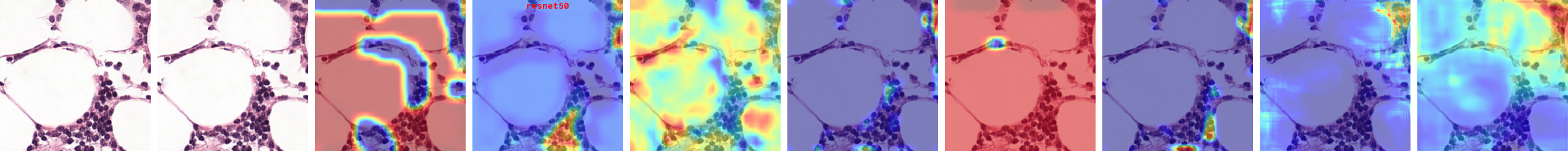}\\
         \includegraphics[width=\linewidth]{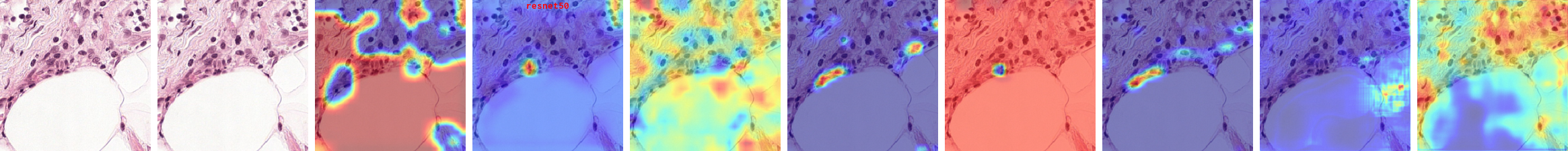}\\
         \includegraphics[width=\linewidth]{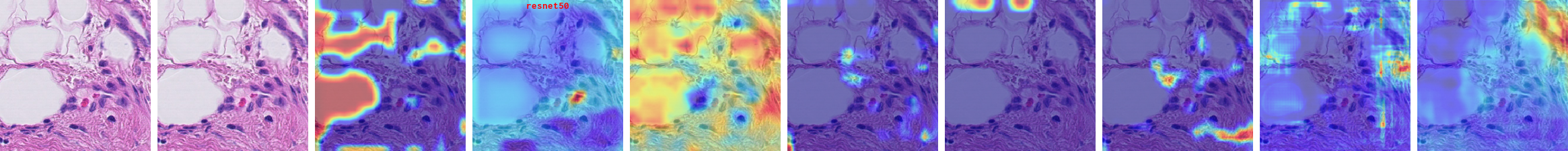}\\
         \includegraphics[width=\linewidth]{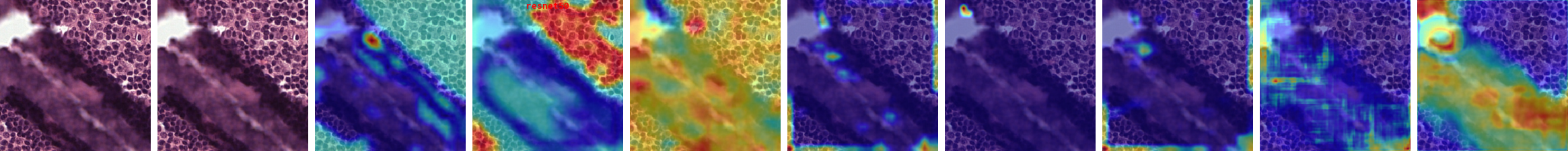}\\
         \includegraphics[width=\linewidth]{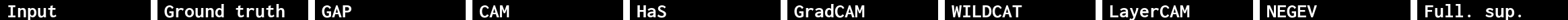}\\
        \caption{Predictions over \textbf{normal} test samples for \camsixteen. Ground truth ROIs are indicated with a red mask highlighting metastatic regions. In all predictions, strong CAM's activations indicate metastatic regions. Backbone: ResNet50.}
        \label{fig:sup-cam16-cams-normal}
\end{figure*}

\begin{figure*}[hbt!]
     \centering
         \centering
         \includegraphics[width=\linewidth]{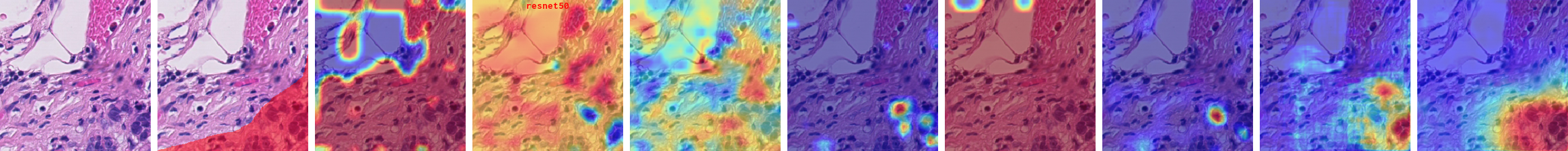}\\
         \includegraphics[width=\linewidth]{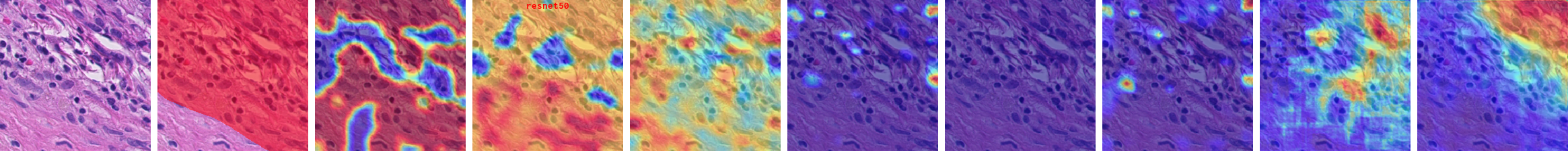}\\
         \includegraphics[width=\linewidth]{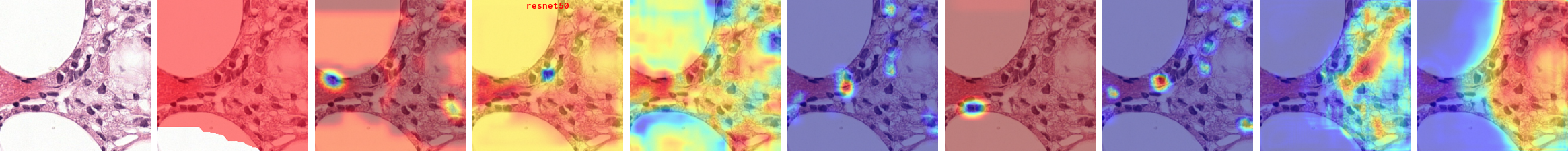}\\
         \includegraphics[width=\linewidth]{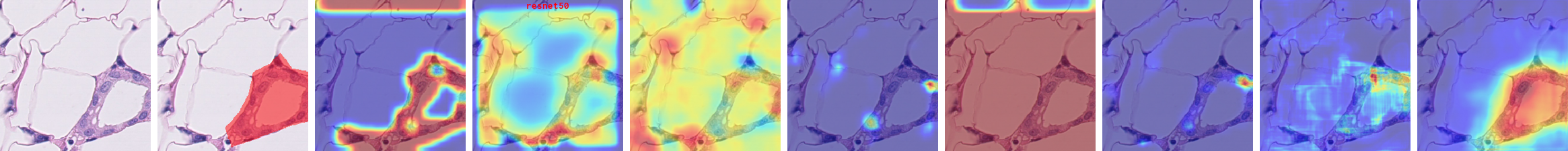}\\
         \includegraphics[width=\linewidth]{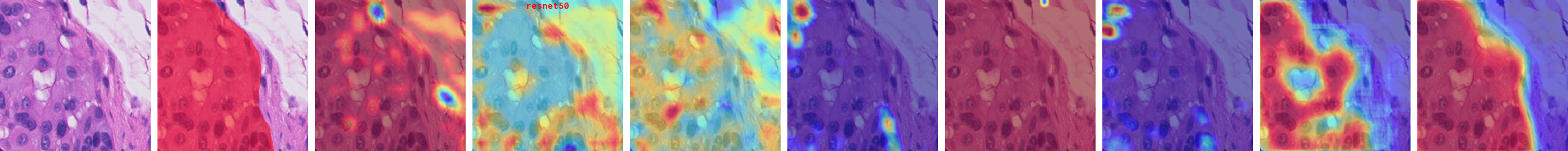}\\
         \includegraphics[width=\linewidth]{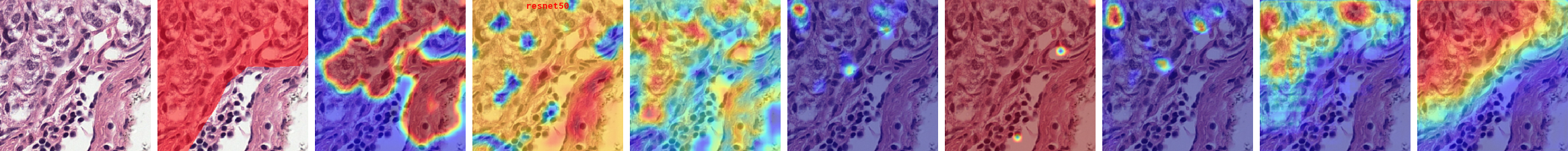}\\
         \includegraphics[width=\linewidth]{tag-CAMELYON512}\\
        \caption{Predictions over \textbf{metastatic} test samples for \camsixteen. Ground truth ROIs are indicated with a red mask highlighting metastatic regions. In all predictions, strong CAM's activations indicate metastatic regions. Backbone: ResNet50.}
        \label{fig:sup-cam16-cams-metastatic}
\end{figure*}

In this section, we provide visual results for the localization of different methods using the ResNet50 backbone: \autoref{fig:sup-glas-cams-benign}, and \autoref{fig:sup-glas-cams-malignant} for \glas test set; and \autoref{fig:sup-cam16-cams-normal} and \autoref{fig:sup-cam16-cams-metastatic} for \camsixteen test set.

\end{document}